\documentclass{article}

\PassOptionsToPackage{numbers, compress}{natbib}
\usepackage[preprint]{neurips_2026}

\usepackage[utf8]{inputenc} 
\usepackage[T1]{fontenc}    
\usepackage{hyperref}       
\usepackage{url}            
\usepackage{booktabs}       
\usepackage{amsfonts}       
\usepackage{nicefrac}       
\usepackage{microtype}      
\usepackage{xcolor}         
\usepackage{graphicx}
\usepackage{enumitem}
\usepackage{amssymb}        
\usepackage{pifont}         
\usepackage{caption}
\usepackage{array}
\usepackage{makecell}
\usepackage{multirow}
\usepackage[table]{xcolor}
\usepackage{adjustbox}
\usepackage{tabularx}
\usepackage{wrapfig}
\usepackage{placeins}
\usepackage{amsmath}
\usepackage{titletoc}
\usepackage{tocloft}

\definecolor{mycolor}{HTML}{E5E5F3} 

\newcolumntype{L}[1]{>{\raggedright\arraybackslash}m{#1}}
\newcolumntype{R}[1]{>{\raggedleft\arraybackslash}m{#1}}

\newcommand{\xmark}{\ding{55}}

\title{AgroTools: A Benchmark for Tool-Augmented Multimodal Agents in Agriculture}

\author{%
\shortstack[c]{%
Zi Ye\textsuperscript{1},
Yibin Wen\textsuperscript{1},
Xiaoya Fan\textsuperscript{2},
Xinyu Zhang\textsuperscript{1},\\
Jing Wu\textsuperscript{1},
Kun Zeng\textsuperscript{1},
Zurong Mai\textsuperscript{1},
Jiarui Zhang\textsuperscript{1},
Bohan Shi\textsuperscript{3},\\
Juepeng Zheng\textsuperscript{1,4}\thanks{Corresponding author. Email: \texttt{zhengjp8@mail.sysu.edu.cn}},
Jianxi Huang\textsuperscript{5,6},
Yutong Lu\textsuperscript{1,4},
Haohuan Fu\textsuperscript{4,7}}\\[0.25em]
\shortstack[c]{%
\textsuperscript{1}Sun Yat-Sen University \quad
\textsuperscript{2}Southwest University \quad
\textsuperscript{3}Northeastern University\\
\textsuperscript{4}National Supercomputing Center in Shenzhen\\
\textsuperscript{5}Southwest Jiaotong University \quad
\textsuperscript{6}China Agricultural University \quad
\textsuperscript{7}Tsinghua University}%
}

\begin{document}

\maketitle

\begin{abstract}
    Agricultural decision-making increasingly requires multimodal systems that can transform visual observations into reliable, executable actions. However, existing agricultural multimodal benchmarks mainly evaluate final-answer correctness and provide limited support for assessing whether models can use external tools to complete precision-sensitive workflows. In this paper, we introduce \textbf{AgroTools}, a benchmark for evaluating tool-augmented multimodal agents in agriculture. AgroTools contains 539 question-answer instances paired with 1,097 heterogeneous agricultural images, spanning five task families and an executable environment of 14 agricultural tools. Each query is annotated with structured tool-use traces, enabling a dual-view evaluation of both process-level execution quality and outcome-level task success. We benchmark 9 open-source and 4 closed-source multimodal large language models on AgroTools. Results show that current models remain far from reliable in agricultural tool-use settings, with clear bottlenecks in tool planning, argument generation, execution recovery, and final-answer synthesis. We hope AgroTools will support future research on multimodal agents for high-precision agricultural applications. 
    The benchmark and evaluation are available at  \url{https://huggingface.co/datasets/AgroTools/AgroTools}.
\end{abstract}

\section{Introduction}

Agriculture is a critical domain for global food security, environmental sustainability, and economic stability~\cite{sporchia2024zero,weiss2020remote,wu2023challenges}. Recent advances in large language models and Multimodal large language models (MLLMs) have opened new opportunities for agricultural applications, such as crop monitoring~\cite{wang2024uas}, pest and disease diagnosis~\cite{wang2025advances}, and decision support~\cite{bie2025adapting,wu2025farmseg_vlm}. In practice, however, many agricultural tasks require more than recognizing visual content or generating a textual answer. A model may need to identify a target crop or pest, retrieve agronomic knowledge, count plants or fruits, measure object sizes, compare multi-temporal observations, and calculate or visualize intermediate results~\cite{gauba2025agmmu}. These requirements make agriculture a natural and important domain for tool-augmented multimodal agents that can combine visual perception, external tools, and step-by-step execution.


\begin{figure}[t]
    \centering
    \includegraphics[width=1\linewidth]{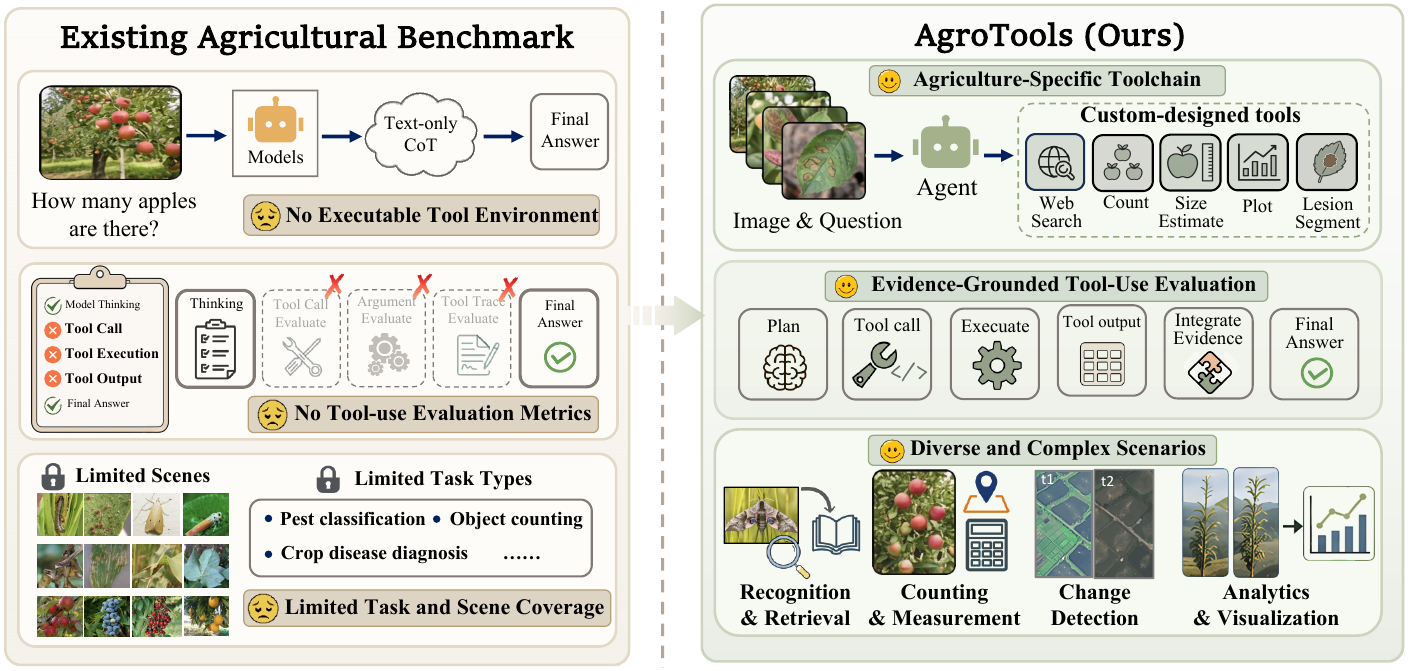}
    \caption{Comparison between existing agricultural benchmarks and AgroTools. Compared with prior benchmarks, AgroTools provides an agriculture-specific toolchain, evidence-grounded tool-use evaluation, and broader coverage of diverse, image-grounded agricultural scenarios.}
    \vspace{-1.5em}
    \label{fig1}
\end{figure}

Recent agricultural multimodal benchmarks have substantially advanced the evaluation of MLLMs in this domain, which incorporate images from web sources, public agricultural datasets, remote sensing platforms, and field-level observations, and construct diverse question-answering tasks~\cite{li2025can, wen2026agrocotchainofthoughtbenchmarkevaluating}.
However, most existing agricultural benchmarks still focus on final-answer correctness. 
In many agricultural tasks, reliable problem solving requires external tools for fine-grained perception, such as numerical computation~\cite{sa2022deepnir} or structured visualization~\cite{sarker2025weedsense}, rather than direct question answering alone. Moreover, final-answer accuracy cannot determine whether a model selects the correct operation, provides valid tool arguments, executes the required procedure, or correctly integrates intermediate outputs. Therefore, evaluating agricultural intelligence requires moving beyond static question answering toward executable tool-use assessment.


In the broader field of language and multimodal agents, tool augmentation has become an important paradigm for solving complex tasks. Models can decompose user instructions, invoke external APIs or tools, observe intermediate results, and iteratively refine their actions through reasoning-acting frameworks such as ReAct~\cite{yao2022react}. A series of benchmarks have been proposed to evaluate general tool-use ability, including API planning and calling~\cite{li-etal-2023-api}, stateful tool interaction~\cite{lu-etal-2025-toolsandbox}, multimodal tool-use planning~\cite{ma2024visual}, and general tool-agent evaluation~\cite{wang2024gta}. These benchmarks show that external tools can extend model capabilities beyond direct generation, especially when tasks require computation, retrieval, or interaction with complex environments.


However, agricultural tool use differs from general tool-use scenarios in several important ways. As illustrated in Fig.~\ref{fig1}, we summarize these differences into three challenges. \textit{(i) Domain-specific execution}: Agricultural tasks often require specialized tools for specific agricultural scenarios such as pest identification, crop disease diagnosis and farmland change detection, rather than generic APIs alone. \textit{(ii) Evidence-grounded synthesis}: agents must not only call the right tools, but also use intermediate observations faithfully to support final answers. Existing tool-use benchmarks, which are mainly designed for open-domain or general multimodal tasks, do not fully capture these agriculture-specific requirements. \textit{(iii) Diversity and complexity of scenarios}: many queries depend on small objects, dense instances, multi-image inputs, or numerical outputs, where minor argument errors can lead to incorrect conclusions. As a result, there remains no unified benchmark for evaluating tool-augmented multimodal agents across diverse, executable, image-grounded agricultural tasks.

To address this gap, we propose \textbf{AgroTools}, an agriculture-specific benchmark for evaluating tool-augmented multimodal agents. AgroTools covers five task families, including recognition and retrieval, counting and measurement, segmentation interpretation, change detection, and visualization and analytics.  Each instance is paired with an executable tool environment consisting of 14 agriculture-oriented tools and annotated with reference tool-use trajectories. We benchmark 13 representative MLLMs under no-tool, step-by-step, and end-to-end settings, revealing systematic bottlenecks in tool planning, argument generation, execution recovery, and answer synthesis.

Overall, the main contributions of this work are summarized as follows:

\begin{itemize}[leftmargin=*]
    \item We introduce \textbf{AgroTools}, a benchmark for evaluating tool-augmented multimodal agents in agriculture. AgroTools contains 539 question-answer instances and 1,097 heterogeneous images from 12 public agricultural datasets. The dataset covers diverse applications such as pest and crop disease diagnosis, segmentation interpretation, and change detection in agricultural scenarios. 
    \item We develop an executable agricultural tool environment with 14 tools tailored to image-grounded agricultural workflows. Each query is paired with a structured reference trajectory, allowing models to be evaluated not only by final-answer correctness, but also by intermediate tool planning, tool selection, argument validity, execution consistency, and final-answer synthesis.
    \item We benchmark 13 open-source and closed-source MLLMs under no-tool, step-by-step, and end-to-end settings. Experiments reveal that current models remain unreliable in agricultural tool-use scenarios, with strong models benefiting more from tool augmentation while weaker models often suffer from long tool instructions, invalid arguments, and poor execution recovery.
    
\end{itemize}

\begin{table*}[t]
  \caption{
    Comparison of representative tool-augmented benchmarks across domain, modality, interaction, and evaluation dimensions.
    \textbf{Dyn.} denotes whether multi-step or interactive tool use is required;
    \textbf{Exec.} denotes whether evaluation is based on actual tool execution or executable outcomes;
    \textbf{Arg.} denotes whether tool arguments are checked;
    \textbf{Traj.} denotes whether intermediate reasoning or tool-use trajectories are evaluated;
    \checkmark: fully supported; \xmark: not supported.
  }
  \vspace{-0.5em}
  \label{tab:tool_eval_comparison}
  \centering
  \small
  \begin{tabular}{lccccccc}
    \toprule
    Benchmark             & Domain           & Multimodal   & Tool         & Dyn.         & Exec.        & Arg.         & Traj.  \\
    \midrule
    MCPTox~\cite{Wang_Gao_Wang_Liu_Sun_Cheng_Shi_Du_Li_2026}                 
                          & General          & \xmark       & \checkmark   & \xmark       & \checkmark   & \xmark       & \xmark \\
    MCP-AgentBench~\cite{Guo_Xu_Zhu_Hong_Wang_Mao_2026}
                          & General          & \xmark       & \checkmark   & \checkmark   & \checkmark   & \xmark       & \xmark \\
    API-Bank~\cite{li-etal-2023-api}
                          & General          & \xmark       & \checkmark   & \checkmark   & \checkmark   & \xmark       & \xmark \\
    ToolSandbox~\cite{lu-etal-2025-toolsandbox}
                          & General          & \xmark       & \checkmark   & \checkmark   & \checkmark   & \checkmark   & \xmark \\
    m\&m’s~\cite{10.1007/978-3-031-72684-2_2}
                          & General          & \checkmark   & \checkmark   & \checkmark   & \checkmark   & \checkmark   & \xmark \\
    GTA~\cite{wang2024gta}
                          & General          & \checkmark   & \checkmark   & \checkmark   & \checkmark   & \checkmark   & \xmark \\
    AgMMU~\cite{gauba2025agmmu}
                          & Agriculture      & \checkmark   & \xmark       & \xmark       & \xmark       & \xmark       & \xmark \\
    AgriMM~\cite{boudiaf2026agrichatmultimodallargelanguage}
                          & Agriculture      & \checkmark   & \xmark       & \xmark       & \xmark       & \xmark       & \xmark \\
    AgroMind~\cite{li2025can}
                          & Agriculture      & \checkmark   & \xmark       & \xmark       & \xmark       & \xmark       & \xmark \\
    AgroCoT~\cite{wen2026agrocotchainofthoughtbenchmarkevaluating}
                          & Agriculture      & \checkmark   & \xmark       & \xmark       & \xmark       & \xmark       & \xmark \\
    
    \midrule
\textbf{AgroTools (Ours)} & Agriculture      & \checkmark   & \checkmark   & \checkmark    & \checkmark  & \checkmark   & \checkmark \\
    
    \bottomrule
  \end{tabular}
  \vspace{-1.5em}
\end{table*}

\vspace{-1em}
\section{Related Work}

\subsection{Agricultural Multimodal Benchmarks}
Multimodal large language models (MLLMs) have shown strong potential in integrating visual perception~\cite{tang2024chain, ma2024visual, he2024multi}, language understanding~\cite{ye2023ureader, zhang2023universal, li2024improving}, and cross-modal reasoning~\cite{fang2025guided, yang2026look, he2025mmboundary, kil2024mllm, chang2026abductivemllm} across real-world domains~\cite{yin2024survey, li2025survey}. This progress has motivated agricultural multimodal models~\cite{sapkota2025multi, eleojo2025utilizing} and benchmarks~\cite{gauba2025agmmu, li2025can, shinoda2025agrobench} for practical agricultural applications. Existing agricultural benchmarks mainly improve data diversity, task coverage, and annotation quality. For example, AgroMind~\cite{li2025can} constructs a multi-dimensional VQA benchmark based on agricultural remote sensing data, while AgMMU~\cite{gauba2025agmmu} introduces real grower--expert dialogues to reflect practical agricultural communication. AgriMM~\cite{boudiaf2026agrichatmultimodallargelanguage} combines visual captioning with web-augmented scientific retrieval grounded in verified phytopathological literature, and benchmarks such as AgriGPT-Omni-2K~\cite{10.1145/3774904.3792971} and AgroBench~\cite{shinoda2025agrobench} further strengthen annotation quality and interaction modality.

Despite these efforts, existing benchmarks focus mainly on answer-centric evaluation and provide limited support for evaluating tool planning, argument correctness, execution fidelity, and trajectory-level reasoning, which leaves a critical gap in assessing the use of external tools for solving dynamic, precision-sensitive agricultural tasks.

\subsection{Tool-Augmented Benchmarks}
Tool-augmented agents extend LLMs and MLLMs from passive response generation to interactive problem solving with external functions, APIs, and domain-specific tools~\cite{zhang2024multimodal, zhu2025agentar, chen2024advancing}. Reasoning--acting paradigms such as ReAct~\cite{yao2022react} and DeepAgent~\cite{10.1145/3774904.3792460} enable models to decompose instructions, invoke tools, observe intermediate results, and refine subsequent actions. These ideas have also inspired domain-specific agent systems~\cite{feng2026earthagentunlockinglandscapeearth}, including MA3~\cite{xu2025multimodalagriculturalagentarchitecture} and AgriDoctor~\cite{11464537}.
Alongside system development, a series of benchmarks have been proposed to evaluate tool-use capabilities~\cite{Guo_Xu_Zhu_Hong_Wang_Mao_2026,Wang_Gao_Wang_Liu_Sun_Cheng_Shi_Du_Li_2026}. API-Bank~\cite{li-etal-2023-api} measures ability to plan, retrieve, and call APIs, while ToolSandbox~\cite{lu-etal-2025-toolsandbox} evaluates intermediate and final milestones along execution trajectories. 
Recent benchmarks have expanded tool-augmented agent evaluation from general multimodal task planning to more realistic settings with user queries, deployed tools, and multimodal inputs~\cite{10.1007/978-3-031-72684-2_2,wang2024gta}, while Earth observation benchmarks further extend this direction to spatial reasoning and EO research through trajectory- and outcome-level assessment~\cite{shabbir2026thinkgeoevaluatingtoolaugmentedagents,feng2026earthagentunlockinglandscapeearth}.


However, these benchmarks often focus on open-domain tasks or Earth observation, lacking the agricultural domain's specific requirements, such as crop knowledge, disease diagnosis and agronomic computations. Furthermore, there is still no dedicated benchmark for assessing how multimodal agricultural agents can reliably use external tools to solve high-precision, real-world tasks. As shown in Table~\ref{tab:tool_eval_comparison}, our benchmark provides comprehensive coverage across domain specificity, dynamic interaction, execution fidelity, argument correctness, and trajectory evaluation.

\vspace{-1em}
\section{Dataset Design}
\vspace{-1em}
\label{Section3}
AgroTools is a problem-oriented, tool-augmented benchmark designed to evaluate multimodal agents on precision-sensitive agricultural tasks. Unlike conventional agricultural multimodal benchmarks that mainly assess final answers or text-only CoT, AgroTools couples each image-grounded query with an executable tool environment, enabling the evaluation of both task outcomes and intermediate tool-use behavior. In this section, we first introduce the task taxonomy of AgroTools and then describe the tool space and interaction protocol that define how agents interact with the benchmark. Benchmark curation, dataset statistics, and evaluation protocols are presented in Sec.~\ref{Section4}.

\vspace{-1em}
\subsection{Task taxonomy} 
\begin{figure}
    \centering
    \includegraphics[width=1\linewidth]{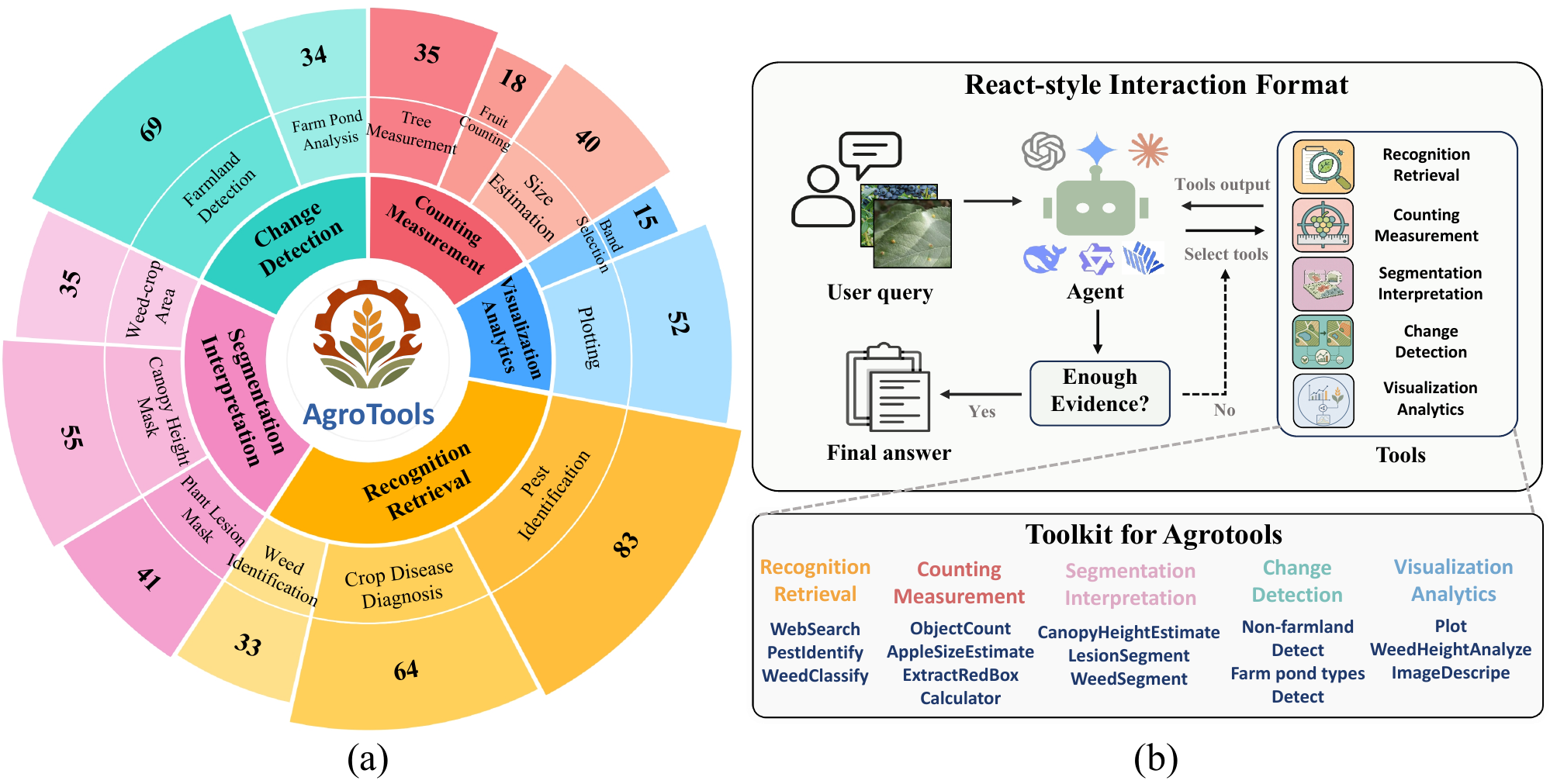}
    \caption{(a) The number of samples across different tasks in AgroTools, some samples can be assigned to more than one task category simultaneously. (b) React-style interaction format and 14 executable tools. The change detection tool involves two scenarios for detecting non-agricultural cultivated land conversion and farmland-pond type changes.}
    \label{fig2}
    \vspace{-2em}
\end{figure}

As illustrated in Fig.~\ref{fig2} (a), AgroTools organizes its instances into five task families: Recognition and Retrieval (\textbf{RR}), Counting and Measurement (\textbf{CM}), Segmentation Interpretation (\textbf{SI}), Change Detection (\textbf{CD}), and Visualization and Analytics (\textbf{VA}). This five-part taxonomy forms a systematic hierarchy of evidence granularity and reasoning complexity in agricultural problem solving. Specifically, it spans from entity-level semantic grounding and instance-level quantitative estimation, to region-level structured interpretation, cross-image or temporal change reasoning, and finally the synthesis of intermediate results into analytical or visual outputs. 

\textbf{Recognition and Retrieval} forms the foundational layer of AgroTools by focusing on the identification of agricultural entities and the retrieval of related agronomic knowledge. This family contains three task types: Pest Identification and Diagnosis, Weed Identification and Retrieval, and Crop Disease Analysis. These tasks require agents to recognize pests, diseases, or plants from visual evidence and connect the recognized target to external agricultural knowledge such as causes, control measures, or species-level information.

\textbf{Counting and Measurement} targets precision-sensitive agricultural tasks that require explicit numerical reasoning over localized objects. This contains three task types: Tree Counting and Measurement, Apple Size Estimation, and Fruit and Crop Organ Counting. These tasks evaluate whether agents can combine visual grounding with quantitative computation to derive counts, sizes, diameters, and other measurement-related outputs from agricultural imagery.

\textbf{Change Detection} focuses on temporally grounded agricultural reasoning across paired observations. This family contains two task types: Farmland Change Detection and Farm Pond Change Analysis. These tasks require agents to determine whether a change has occurred, identify the corresponding transition type, and quantify changed regions or proportions from multi-temporal agricultural imagery.

\textbf{Segmentation Interpretation} focuses on tasks in which the final answer depends on interpreting structured masks or region-level visual evidence. This family contains three task types: Lesion Mask \& Severity Analysis, Canopy Height Mask Interpretation, and Weed–Crop Area Interpretation. These tasks assess whether agents can transform pixel-level or mask-level outputs into agronomically meaningful judgments, such as severity estimation and vegetation coverage analysis.

\textbf{Visualization and Analytics} emphasizes structured reasoning over quantitative evidence and intermediate results, which contains two task types: Band-Combination Selection and Plotting Tasks. These tasks assess whether agents can compare alternatives, organize intermediate values, rank candidate settings, and generate explicit visual summaries such as curves or charts from agricultural inputs.

\vspace{-1em}
\subsection{Tool Suite and Task Format} 
\vspace{-0.5em}
\textbf{Task Format.} 
Each AgroTools instance is formulated as a step-implicit and tool-implicit query under a ReAct-style interaction format (see Fig.~\ref{fig2} (b)). Given an input image and a natural language instruction, the agent is required to infer the necessary intermediate steps, select appropriate tools from the predefined toolbox, generate valid tool arguments, and produce the final answer based on intermediate observations. This design avoids exposing explicit procedural hints or tool names in the prompt, thereby better reflecting realistic workflows in which multimodal agents must determine both what to do and how to do it.

\textbf{Tool Space.}
AgroTools provides 14 executable tools that span the major computational capabilities required by the benchmark (see Fig.~\ref{fig2} (b)). For clarity, we organize them into five functional categories: (i) identification and retrieval tools, which support pest and crop disease diagnosis, weed species recognition, and agronomic knowledge lookup; (ii) localization and quantitative estimation tools, which support marked-region extraction, object counting, fruit-size estimation, and weed phenotyping; (iii) segmentation and spatial analysis tools, which support lesion analysis, region-level interpretation, canopy height estimation, and crop–weed area analysis; (iv) temporal change-analysis tools, which support transition detection and quantification across paired observations; and (v) analytical and visualization utilities, which support numerical calculation and plot generation. Together, these tools produce diverse outputs, including class labels, bounding boxes, counts, diameters, phenotypic traits, pixel-level statistics, change masks, and figures. Detailed tool specifications are provided in Appendix~\ref{appendix:tool details}.

\begin{figure}
    \centering
    \includegraphics[width=0.91\linewidth]{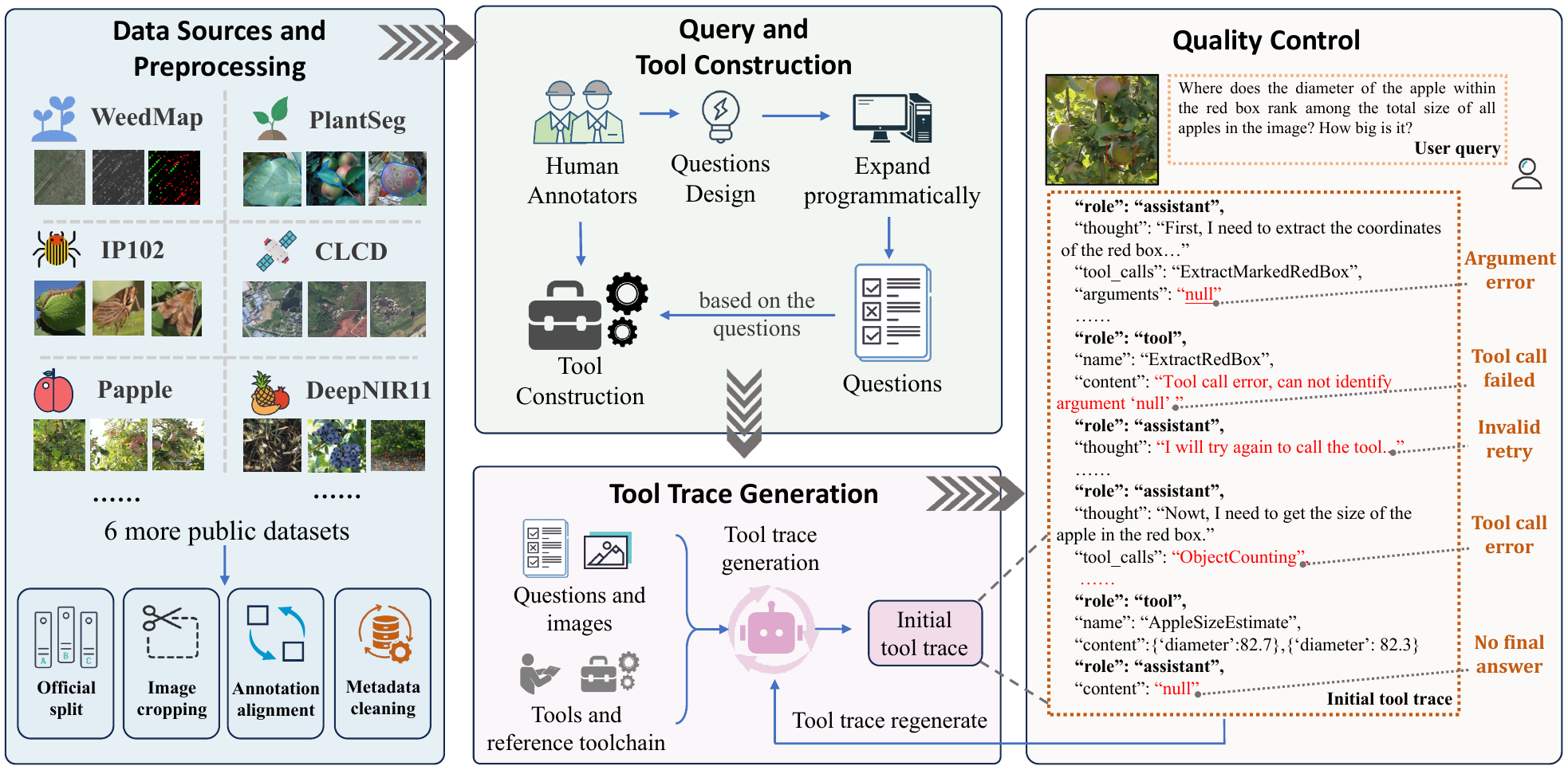}
    \caption{AgroTools curation pipeline contains four stages: Data Sources and Preprocessing, Query and Tool Construction, Tool Trace Generation, and Quality Control.}
    \label{fig3}
    \vspace{-2em}
\end{figure}

\begin{figure*}[t]
    \centering

    \begin{minipage}[t]{0.37\textwidth}
    \vspace*{4mm}
    \centering
    \small
    \setlength{\tabcolsep}{4pt}
    \renewcommand{\arraystretch}{1.08}

    \begin{tabular}{L{2.8cm} R{1.6cm}}
        \toprule
        Item & Number \\
        \midrule
        Total queries & 539 \\
        Text-only answers & 487 \\
        Plot outputs & 52 \\
        Unique image files & 1097 \\
        \midrule
        Total tool calls & 1447 \\
        \makecell[l]{Average tool calls\\per query} & 2.68 \\
        \makecell[l]{1/2/3/4-tool\\queries} & \makecell[r]{59/412\\49/19} \\
        \bottomrule
    \end{tabular}

    \vspace{2mm}
    \captionsetup{type=table, width=\linewidth}
    \caption{Overview of AgroTools benchmark scale and annotation complexity, summarizing query volume, image resources, answer modalities, and tool-call composition.}
    \label{tab2}
    \end{minipage}
    \hfill
    \begin{minipage}[t]{0.6\textwidth}
        \vspace{0pt}
        \centering
        \includegraphics[width=\linewidth]{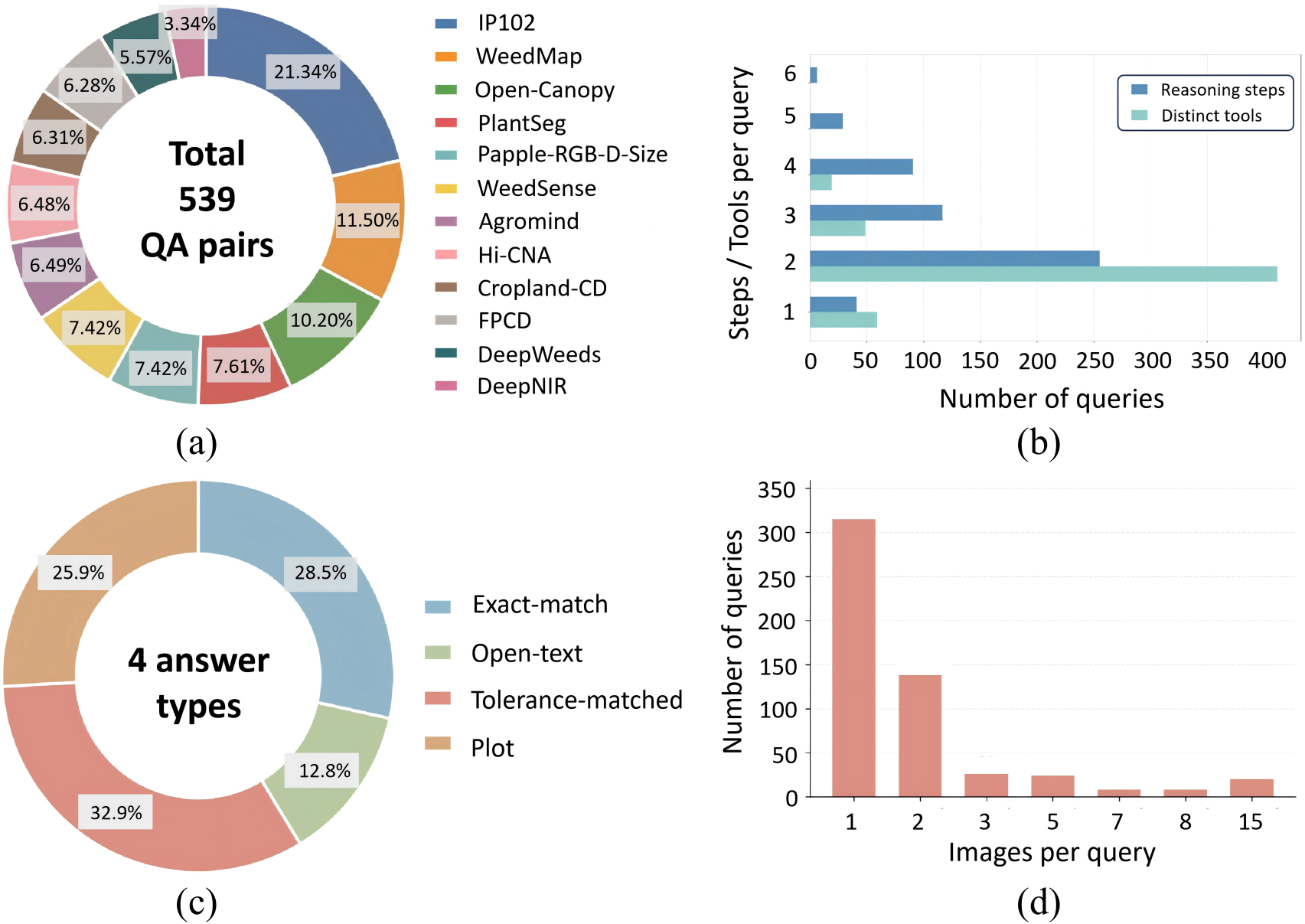}
        \captionsetup{type=figure, width=\linewidth}
        \caption{Detailed statistics of AgroTools.}
        \label{fig4}
    \end{minipage}
    \vspace{-1em}
\end{figure*}

\section{AgroTools Benchmark}
\label{Section4}

\subsection{Benchmark curation}
\textbf{Data Sources and Preprocessing.} As shown in Fig.~\ref{fig3}, our benchmark is constructed from 12 public agricultural datasets, including WeedMap~\cite{sa2018weedmap}, Open-Canopy~\cite{fogel2025open}, PlantSeg~\cite{wei2026large}, PApple RGB‑D Size dataset~\cite{ferrer2023simultaneous}, WeedSense~\cite{sarker2025weedsense}, Hi-CNA~\cite{sun2024identifying}, AgroMind~\cite{li2025can}, CropLand-CD~\cite{liu2022cnn}, FPCD~\cite{tundia2023fpcd}, DeepWeeds~\cite{olsen2019deepweeds}, deepNIR~\cite{sa2022deepnir}, and IP102~\cite{wu2019ip102}. These datasets collectively cover diverse agricultural task types. For example, Open-Canopy supports canopy height regression and vegetation cover area segmentation, Hi-CNA supports non-farmland change detection, and deepNIR supports fruit counting. We follow the official split of each dataset whenever available. For ultra-high-resolution imagery, such as WeedMap, we adopt the official cropping strategy and naming convention to preserve consistency with the original data organization. After preprocessing, we further align each image with its corresponding annotation and remove samples with incomplete metadata.

\textbf{Query and Tool Construction.}
We first reviewed prior agricultural benchmarks and application studies to identify representative and practically meaningful question types in agricultural scenarios, then we recruited human volunteers to design agriculture-oriented questions grounded in the images and supported by the source annotations. These seed questions were then expanded programmatically through templates to improve coverage while preserving task consistency. Based on the resulting queries, we constructed the corresponding tools 
and verified their executability. To reduce evaluation ambiguity, we filtered queries to retain cases with a unique and well-defined tool invocation order. 

\textbf{Gold Trace Generation.} For each query, human annotators first specified a reference sequence of tool invocations as the target solution path. Based on this reference order, we then used GPT-4o~\cite{hurst2024gpt} to generate an initial tool trace containing tool calls, arguments, intermediate outputs, and a preliminary answer. This procedure reduces annotation cost while keeping the core reasoning process under human control, and also ensures a unified trace format across different task types.

\textbf{Quality Control.} All generated traces were manually reviewed before inclusion in the benchmark. We checked whether the tool order matched the reference sequence, whether the arguments were correct, and whether the tool chain was complete and executable. For open-ended tasks, we further verified that the returned answers were consistent with established agronomic knowledge and objective facts. For plotting and other generation tasks, we examined whether the input parameters were faithful to tool outputs and whether the generated results satisfied the task requirements.

\subsection{Statistics}
Basic benchmark statistics are presented in Table~\ref{tab2}. The benchmark involves a total of 539 queries, 1,097 unique image files, 14 executable tools, 487 queries with text-only answers and 52 queries with plot outputs. The annotated tool-use traces contain 1,447 tool calls in total, with an average of 2.68 tool calls per query. The number of tools involved in each query varies from 1 to 4, with most queries requiring 2 to 4 reasoning steps (see Fig.~\ref{fig4}(b)). 
As shown in Fig.~\ref{fig4}(a), AgroTools was curated from 12 agricultural data sources. IP102 contributes the largest proportion (21.34\%), followed by WeedMap (11.50\%) and Open-Canopy (10.20\%), while the remaining sources each account for between 3.34\% and 7.61\%. This distribution shows that AgroTools is constructed from diverse agricultural benchmarks rather than being dominated by a single source. AgroTools also covers four answer types (see Fig.~\ref{fig4}(c)), including tolerance-matched (correct within allowable error range), exact-match, plot, and open-text answers, accounting for 32.9\%, 28.5\%, 25.9\%, and 12.8\%, respectively. In addition, most queries are grounded in one or two images, although a subset involves larger image sets, indicating AgroTools also includes complex multi-image scenarios (see Fig.~\ref{fig4}(d)).

\vspace{-1em}
\subsection{Evaluation protocol}
\vspace{-0.5em}
\textbf{Evaluation modes.} We report results under three evaluation modes. \textbf{No-tool} mode evaluates the model's native multimodal reasoning ability without any external tools. In this setting, the model is given only the question and the associated image(s), and is required to produce the answer directly. For queries involving raster-format image processing and plotting, the required operations cannot be completed without tool support. We therefore exclude these queries from no-tool evaluation and report results only on tasks that can be solved directly from the visual input and question. \textbf{Step-by-step} mode evaluates fine-grained tool-use ability. In this setting, the model is provided with the preceding ground-truth interaction history from the reference trajectory and is asked to predict the next action or response at each step. This mode enables direct comparison with the reference trajectory while reducing the impact of error accumulation. \textbf{End-to-end} mode evaluates the model as a complete agent in an executable environment. The model must autonomously plan, invoke tools, interpret intermediate outputs, and produce final answers. Because each decision depends on actual outcomes of previous steps, this mode better reflects realistic task execution and overall agent performance.

\textbf{Evaluation metrics.} 
We adopt a dual-view metric suite that evaluates both process-level tool-use behavior and outcome-level task success. Under step-by-step mode, we report five process-level metrics: Step-Type Accuracy (\textbf{\textit{STA}}), Tool Accuracy (\textbf{\textit{ToolAcc}}), Argument Accuracy (\textbf{\textit{ArgAcc}}), Early Answer Rate (\textbf{\textit{EAR}}), and Summary Accuracy (\textbf{\textit{SummAcc}}). STAcc measures whether the model predicts the correct type of the next step in the reference trajectory, such as issuing a tool call rather than directly producing a final answer. ToolAcc measures whether the selected tool matches the reference tool whenever a tool invocation is required. ArgAcc measures whether the generated tool arguments are valid and consistent with the reference invocation. EAR measures the proportion of intermediate steps at which the model outputs a final answer before the required tool-use procedure is completed. SummAcc evaluates whether the model can produce the correct final answer when conditioned on the preceding ground-truth interaction history.

For no-tool mode and end-to-end mode, we evaluate final-answer quality using a unified set of outcome-level metrics: Final Answer Score (\textbf{\textit{FAS}}), Closed-Slot Score (\textbf{\textit{CSS}}), and Open-Text Score (\textbf{\textit{OTS}}). Final Answer Score serves as the overall outcome-level metric by averaging the type-appropriate score for each query, thereby providing a unified measure of final-answer quality across different answer formats. Closed-Slot Score is used for questions with tightly constrained answers, corresponding to the exact-match and tolerance-matched answer types shown in Fig.~\ref{fig4}(c). Open-Text Score evaluates open-text questions that allow flexible natural-language responses. In addition, we also scored the plot-type tasks. Detailed definitions of all metrics are provided in Appendix~\ref{appendix:evaluation metrics}.



\vspace{-1em}
\section{Experiments}

\begin{table*}[t]
\centering
\caption{Main results across three evaluation modes. Tool., Arg., Summ. denote ToolAcc, ArgAcc, and SummAcc respectively. Best results across all models are shown in \textbf{bold}, and second-best results are \underline{underlined}. For fair comparison, no-tool and end-to-end results in this table are reported on the shared subset that excludes raster-format image-processing and plot-generation tasks. }
\vspace{-0.5em}
\label{tab3:main_results}

\scriptsize
\setlength{\tabcolsep}{3.2pt}
\setlength{\arrayrulewidth}{0.35pt}
\renewcommand{\arraystretch}{1.10}

\begin{adjustbox}{max width=\linewidth}
\begin{tabular}{l|ccccc|ccc|ccc}
\toprule
\multirow{2}{*}{\textbf{Model}}
& \multicolumn{5}{c|}{\textbf{Step-by-step Mode}}
& \multicolumn{3}{c|}{\textbf{No-tool Mode}}
& \multicolumn{3}{c}{\textbf{End-to-end Mode}} \\
\cmidrule(lr){2-6}\cmidrule(lr){7-9}\cmidrule(lr){10-12}
& \textbf{STA} $\uparrow$
& \textbf{Tool.} $\uparrow$
& \textbf{Arg.} $\uparrow$
& \textbf{EAR} $\downarrow$
& \textbf{Summ.} $\uparrow$
& \textbf{CSS} $\uparrow$
& \textbf{OTS} $\uparrow$
& \textbf{FAS} $\uparrow$
& \textbf{CSS} $\uparrow$
& \textbf{OTS} $\uparrow$
& \textbf{FAS} $\uparrow$ \\
\midrule

\rowcolor{mycolor}
\multicolumn{12}{l}{\textbf{\textit{Closed-source}}} \\

GPT-5.4~\cite{leon2025gpt}
& 65.86 & 52.32 & 26.26 & 46.79 & 10.93
& \underline{24.83} & \textbf{33.81} & \textbf{24.51}
& \textbf{63.35} & \underline{65.20} & \textbf{62.74} \\

Claude-Sonnet-4.6~\cite{anthropic2026claudesonnet46}
& 67.17 & 50.79 & 23.15 & 45.06 & 25.53
& 20.83 & \underline{29.09} & 20.97
& 21.86 & 33.79 & 25.44 \\

Gemini-2.5-Pro~\cite{comanici2025gemini}
& 97.18 & \underline{91.91} & \underline{47.06} & 3.59 & 21.01
& 21.53 & 10.46 & 17.45
& 39.26 & 42.77 & 40.68 \\

Doubao-Seed-2.0-Pro~\cite{seed2026doubao}
& \underline{98.44} & \textbf{95.16} & \textbf{49.34} & 2.07 & 17.52
& 22.89 & 21.92 & 18.92
& \underline{48.32} & 63.85 & 48.64 \\

\midrule

\rowcolor{green!12}
\multicolumn{12}{l}{\textbf{\textit{Open-source}}} \\

InternVL3-8B~\cite{chen2024expanding}
& 51.76 & 20.66 & 5.32 & 63.37 & 20.52
& 12.85 & 10.79 & 10.62
& 2.13 & 1.74 & 1.61 \\

InternVL3.5-8B~\cite{chen2024expanding}
& 96.43 & 85.49 & 31.58 & 2.49 & \underline{33.72}
& 13.89 & 12.73 & 13.62
& 32.31 & 46.40 & 36.18 \\

InternVL3.5-14B~\cite{chen2024expanding}
& 94.41 & 85.00 & 32.27 & 7.67 & \textbf{33.96}
& 15.10 & 19.88 & 14.28
& 38.56 & 51.50 & 43.74 \\

Qwen3.5-9B~\cite{bai2023qwen}
& 95.57 & 86.66 & 27.44 & 5.25 & 16.70
& \textbf{25.35} & 27.39 & \underline{23.13}
& 40.41 & 59.05 & 39.19 \\

Qwen3.5-35B-A3B~\cite{bai2023qwen}
& 96.58 & 91.71 & 40.70 & 3.32 & 13.88
& 16.15 & 12.84 & 13.84
& 32.09 & 59.91 & 38.53 \\

Qwen3-VL-8B-Instruct~\cite{bai2023qwen}
& 83.23 & 67.31 & 23.43 & 23.01 & 28.24
& 21.88 & 16.48 & 17.23
& 45.92 & 52.10 & 41.64 \\

LLaVA-NeXT-13B~\cite{liu2024improved}
& \textbf{99.80} & 44.85 & 5.46 & \underline{0.07} & 21.83
& 5.90 & 15.60 & 6.82
& 0.69 & 1.87 & 0.71 \\

Janus-Pro-7B~\cite{chen2025janus}
& 76.59 & 15.96 & 0.21 & \textbf{0.00} & 1.34
& 14.06 & 23.92 & 13.88
& 1.22 & 2.69 & 1.86 \\

GLM-4.6V~\cite{hong2025glm}
& 87.71 & 84.93 & 39.53 & 9.47 & 22.48
& 16.15 & 14.35 & 13.87
& 46.51 & \textbf{68.94} & \underline{52.15} \\

\bottomrule
\end{tabular}
\end{adjustbox}
\vspace{-1.5em}
\end{table*}

\subsection{Experiment settings}
We evaluate 13 multimodal models on AgroTools, including four proprietary models and nine open-source models. The proprietary models are GPT-5.4~\cite{leon2025gpt}, Claude Sonnet 4.6~\cite{anthropic2026claudesonnet46}, Gemini 2.5 Pro~\cite{comanici2025gemini}, and Doubao Seed 2.0 Pro~\cite{seed2026doubao}. The open-source models include InternVL series~\cite{chen2024expanding}, Qwen series~\cite{bai2023qwen}, LLaVA-v1.6-Vicuna-13B~\cite{liu2024improved}, GLM-4.6V~\cite{hong2025glm}, and Janus-Pro-7B~\cite{chen2025janus}. Experiments are conducted on a server equipped with an NVIDIA A800 80GB GPU. All evaluations are conducted in a zero-shot setting with standardized decoding configurations for fair comparison across models, and detailed experimental settings are provided in Appendix~\ref{appendix:experiment details}.

\begin{table*}[t]
\centering
\caption{Error breakdown for the two most challenging task categories. ParamErr indicates that at least one tool call in the corresponding toolchain contains an invalid parameter or format error. SwitchErr denotes cases where the model switches to an incorrect tool after a failed tool call, and Unsup.Ans. denotes answers that are not sufficiently supported by tool outputs. Halluc. denotes fabricated plot contents or numerical trends without sufficient tool evidence, and PlotFail denotes failures to produce a valid final plot output.}
\vspace{-0.5em}
\label{table4}
\small
\setlength{\tabcolsep}{5pt}
\renewcommand{\arraystretch}{1.12}
\begin{adjustbox}{max width=\linewidth}
\begin{tabular}{l|ccc|ccc}
\toprule
\multirow{2}{*}{\textbf{Model}}
& \multicolumn{3}{c|}{\textbf{Change Detection}}
& \multicolumn{3}{c}{\textbf{Visualization \& Analytics}} \\
\cmidrule(lr){2-4}\cmidrule(lr){5-7}
& \textbf{ParamErr}
& \textbf{SwitchErr}
& \textbf{Unsup.Ans.}
& \textbf{ParamErr}
& \textbf{Halluc.}
& \textbf{PlotFail} \\
\midrule
GPT-5.4 &  \textbf{56.3} & 7.8 & \textbf{49.5} & \textbf{9.6} & 7.7 & \textbf{19.2} \\
Doubao-Seed-2.0-Pro & 83.5 & \textbf{3.9} & 79.6 & 76.9 & \textbf{3.8} & 28.8 \\
GLM-4.6V & 70.9 & 64.1 & 54.4 & 98.1 & 55.8 & 61.5 \\
Qwen3.5-35B-A3B & 73.8 & 63.1 & 67.0 & 100.0 & 51.9 & 44.2 \\
\bottomrule
\end{tabular}
\end{adjustbox}
\vspace{-2em}
\end{table*}

\begin{figure}[t]
    \centering
    \includegraphics[width=1\linewidth]{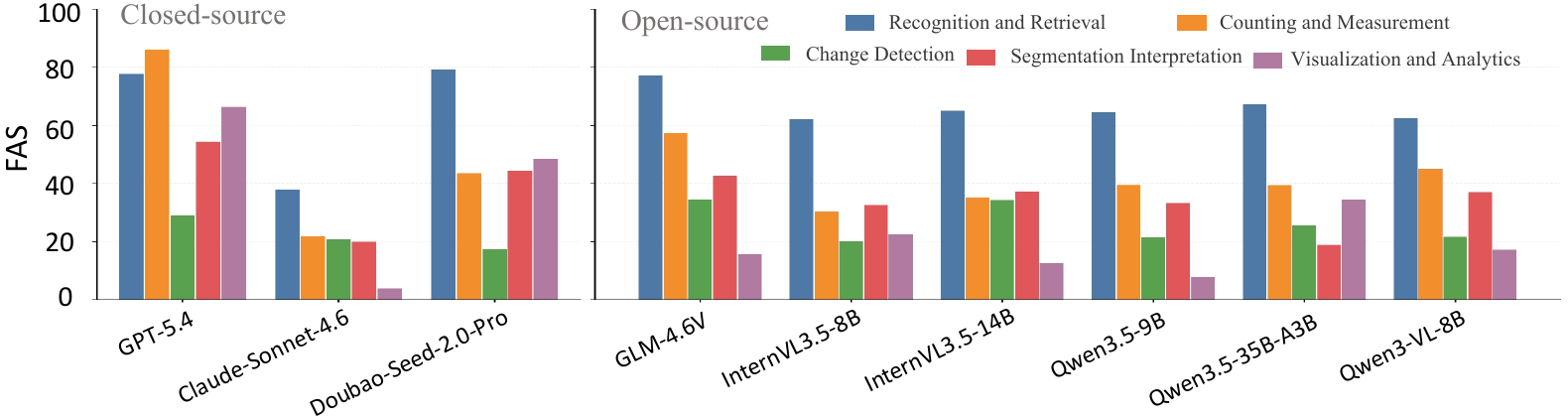}
    \caption{Task-level FAS across representative models. Scores are computed over all samples in the full dataset within each task category; plot-generation samples are scored by the plot evaluator.}
    \label{fig5}
    \vspace{-2em}
\end{figure}


\subsection{Main results}

\textbf{Tool augmentation shows mixed effects.} Table~\ref{tab3:main_results} shows that the effect of end-to-end tool use is positive for most models. On the shared subset, compared with no-tool mode, end-to-end tool use improves CSS, OTS, and FAS for 10 out of 13 models, while the remaining 3 models show declines on all three metrics. The clearest gains are mostly observed on more capable models with stronger long-context understanding and instruction-following ability. For example, GPT-5.4 and Doubao-Seed-2.0-Pro achieve large FAS improvements after tool augmentation, and some stronger open-source models, such as Qwen3-VL-8B-Instruct and GLM-4.6V, also benefit substantially. This suggests that external tools can help models solve agricultural tasks that require retrieval, counting, calculation, and multi-step reasoning. However, several weaker open-source models, such as Janus-Pro-7B and LLaVA-NeXT-13B, perform significantly worse after tool augmentation, with FAS dropping from 13.88 to 1.86 and from 6.82 to 0.71, respectively. In the end-to-end setting, the ReAct-style interaction introduces long tool instructions and specific model output requirements, which likely further degrades their output quality.

\textbf{Step-by-step mode analysis.} Table~\ref{tab3:main_results} reveals a consistent gap between coarse action prediction and precise tool execution. Across models, STAcc and ToolAcc are generally much higher than ArgAcc, indicating that many models can identify the correct next step and even select the appropriate tool, yet still fail to generate arguments that are valid and faithful to the required operation. For example, Gemini-2.5-Pro drops from 91.91 ToolAcc to 47.06 ArgAcc, and Doubao-Seed-2.0-Pro from 95.16 to 49.34. EAR further reveals different failure tendencies across model families. Closed-source models such as GPT-5.4 (46.79) and Claude-Sonnet-4.6 (45.06) exhibit notably high EAR, suggesting a tendency to terminate the procedure prematurely, whereas most open-source models follow the required tool sequence more faithfully. However, low EAR alone does not guarantee correct final outputs, as SummAcc remains modest across all models (at most 33.96), indicating that integrating intermediate tool outputs into a correct final answer is itself a challenging capability.

\subsection{Discussion}
\textbf{Performance varies substantially across task types.} Fig.~\ref{fig5} reports representative models selected to cover proprietary and open-source families and different overall performance levels. In particular, recognition and retrieval is relatively stable across stronger models, whereas larger variations emerge on change detection, segmentation interpretation, and especially statistical analysis. For example, GPT-5.4 remains strong on detection, counting, and measurement (86.03) and also performs best on statistical analysis (66.37), but its score drops sharply on change detection (29.03), indicating that even top-performing models still struggle with temporally grounded comparison and cross-image reasoning. A similar heterogeneity also appears in open-source models, GLM-4.6V achieves competitive results on recognition and retrieval (77.34), change detection (34.47), yet falls to 15.77 on statistical analysis, while Qwen3.5-35B-A3B performs relatively better on statistical analysis (34.38) but remains weak on segmentation interpretation (18.92).

\textbf{Analysis of failure patterns in weak categories.} Since Change Detection and Visualization and Analytics show consistently low or highly unstable FAS across representative models, we further analyze their intermediate-process failures and summarize the most frequent error types in Table~\ref{table4}. The results suggest that these categories are mainly limited by execution reliability and evidence grounding. In Change Detection, parameter errors are frequent for all representative models, ranging from 56.3 for GPT-5.4 to 83.5 for Doubao-Seed-2.0-Pro, showing that models often fail to correctly specify paired-image inputs and change-analysis arguments. GLM-4.6V and Qwen3.5-35B-A3B also show high SwitchErr rates above 63, indicating that they tend to deviate from the expected toolchain after a failed call rather than repair the original action. In Visualization and Analytics, open-source models suffer more from parameter failures, with GLM-4.6V reaching 98.1 and Qwen3.5-35B-A3B reaching 100.0, and their high Halluc. and PlotFail rates show that plot-generation tasks require faithful transfer of intermediate numerical evidence into the final output. These errors show that weak categories require robust tool execution, error recovery, and grounded synthesis across multiple intermediate steps. For more error case analyses, see Appendix~\ref{appendix:case study}.


\textbf{Models differ in execution recovery patterns.}
Fig.~\ref{fig:tool error} summarizes successful tool calls, failed calls followed by retries, and failed calls without retry in the end-to-end setting. Several models exceed the 1,447 reference tool calls, showing that they often enter repeated execution cycles. However, more failed calls do not necessarily indicate weaker tool use, since they may also reflect a certain degree of self-correction after execution errors, such as retrying with revised arguments after an invalid call. For example, InternVL3.5-14B and InternVL3.5-8B produce many failed-but-retried calls, suggesting that they can continue interacting with the tools, although their repairs are not always effective. In contrast, Claude-Sonnet-4.6 has fewer failed calls but almost no retries, indicating limited recovery once a tool call fails. Stronger models such as GPT-5.4 and GLM-4.6V achieve a better balance between successful calls and unrecovered failures, partly explaining their stronger end-to-end FAS in Table~\ref{tab3:main_results}. These results suggest that reliable agricultural agents require not only correct initial tool use, but also execution-aware retry and repair.

\section{Conclusion}
We introduce AgroTools, an agriculture-specific benchmark for evaluating tool-augmented multimodal agents. AgroTools pairs image-grounded agricultural queries with executable tools and structured reference trajectories, enabling an evaluation of both final task success and intermediate tool-use behavior. Covering 539 queries, 1,097 images, five task families, and 14 agriculture-oriented tools, AgroTools provides a unified testbed for precision-sensitive agricultural workflows. Experiments with 13 representative MLLMs show that current models are still far from reliable agents in agricultural scenarios, with persistent failures in tool planning, argument generation, execution recovery, and answer synthesis. We hope AgroTools will support future progress toward trustworthy multimodal agents for real-world agricultural applications.



\bibliographystyle{plainnat}
\bibliography{references}


\clearpage
\appendix

\newpage
\appendix
\begin{center}
    \Large
    \textbf{AgroTools: A Benchmark for Tool-Augmented Multimodal Agents in Agriculture \\(Supplementary material)} 
\end{center}


\setlength{\cftbeforesecskip}{1pt}
\setlength{\cftbeforesubsecskip}{0pt}
\setlength{\cftbeforesubsubsecskip}{0pt}
\setlength{\cftaftertoctitleskip}{1pt}

\startcontents[sections]
\printcontents[sections]{}{1}{\section*{Table of Contents in Appendix}\setcounter{tocdepth}{3}}

\newpage

\newpage
\section{Broader Impacts}
Tool-augmented multimodal agents have growing potential in agriculture, where many real-world tasks require combining visual observations with external knowledge, numerical analysis, and structured decision support. By introducing AgroTools, we aim to provide a standardized and executable benchmark for evaluating whether such systems can support precision-sensitive agricultural workflows more reliably. We hope this benchmark will facilitate research on trustworthy agricultural AI, promote reproducible comparison across models, and help the community identify failure modes before these systems are considered for high-impact applications such as crop monitoring, pest and disease analysis, land-use assessment, and resource management.

AgroTools may also contribute positively by encouraging evaluation beyond answer-only correctness. In agriculture, a seemingly plausible final answer can still be unsupported by the underlying evidence or produced through an unreliable procedure. By making tool planning, argument generation, execution behavior, and evidence-grounded synthesis observable and measurable, the benchmark can support the development of models that are more transparent, better calibrated, and more useful for human-in-the-loop agricultural decision support.

At the same time, there are potential risks. Benchmarking advances in tool-augmented agents may accelerate interest in deploying such systems in real agricultural workflows, and incorrect outputs could lead to inappropriate recommendations if used without expert oversight. Dataset coverage may also remain uneven across crops, regions, or management contexts, which can introduce performance gaps in underrepresented scenarios. For this reason, AgroTools is intended as a research benchmark rather than an autonomous decision-making system. All underlying data are drawn from publicly available sources, and we plan to release the benchmark and evaluation code in accordance with the licenses and terms of the original datasets. We encourage future use of the benchmark in settings that retain agronomic expert oversight and task-specific verification.

\section{Data Collection}

\subsection{Data Sources}
AgroTools is constructed based on twelve publicly available agricultural vision-language datasets (see Table~\ref{tab1:dataset summary}), each offering unique characteristics and specialized annotations for comprehensive evaluation of tool-augmented multimodal agents in agriculture.

\paragraph{WeedMap~\cite{sa2018weedmap}} enhances our benchmark with large-scale aerial multispectral imagery for semantic weed mapping. The dataset comprises high-resolution imagery collected over agricultural fields, with dense annotations for weed species localization and classification. This dataset supports weed identification and retrieval tasks in precision farming scenarios, enabling evaluation of agents' ability to recognize and localize weed species from aerial observations.

\paragraph{Open-Canopy~\cite{fogel2025open}} contributes very high-resolution forest monitoring data, enabling canopy height regression and vegetation cover area segmentation. The dataset combines RGB imagery with detailed canopy structure annotations, supporting segmentation interpretation tasks that require transforming pixel-level outputs into agronomically meaningful judgments such as vegetation coverage analysis and height estimation.

\paragraph{PlantSeg~\cite{wei2026large}} provides a large-scale in-the-wild dataset for plant disease segmentation. The dataset contains diverse disease images captured under real field conditions, with pixel-level lesion masks and severity annotations. This dataset supports lesion mask analysis and severity estimation, evaluating agents' capability to interpret structured mask outputs for disease diagnosis.

\paragraph{PApple RGB‑D Size dataset~\cite{ferrer2023simultaneous}} offers RGB-D images of apples with precise size annotations. The dataset combines visual imagery with depth information, enabling fruit detection, diameter estimation, and size measurement tasks. This dataset supports counting and measurement tasks that require explicit numerical reasoning over localized objects in agricultural imagery.

\paragraph{WeedSense~\cite{sarker2025weedsense}} provides multi-task learning annotations for comprehensive weed analysis. The dataset supports weed segmentation, height estimation, and growth stage classification through densely annotated RGB images. This dataset enables evaluation of agents' ability to perform fine-grained weed phenotyping and integrate multiple visual understanding tasks into coherent tool-use trajectories.

\paragraph{Hi-CNA~\cite{sun2024identifying}} focuses on cropland non-agriculturalization change detection using bi-temporal high-resolution remote sensing images. The dataset provides paired observations with change annotations, supporting the identification of farmland-to-non-farmland transitions. This dataset evaluates agents' capability for temporally grounded agricultural reasoning, including transition detection and change area quantification.

\paragraph{AgroMind~\cite{li2025can}} serves as a comprehensive agricultural remote sensing benchmark within our integrated dataset. It comprises 28,482 question-answer pairs and 20,850 images sourced from nine public datasets and one proprietary global parcel dataset, spanning diverse agricultural scenarios including pest and disease monitoring, parcel analysis, multiple crop recognition, and anomaly detection. The dataset uniquely covers four hierarchical task dimensions—spatial perception, object understanding, scene understanding, and scene reasoning—with 13 specific task types ranging from basic classification to advanced reasoning, integrating multi-sensor data from UAVs, satellites, and ground cameras.

\paragraph{CropLand-CD~\cite{liu2022cnn}} provides fine-grained cropland change detection annotations using CNN-transformer networks. The dataset focuses on detecting subtle changes in agricultural land use over time, supporting temporal change analysis tasks that require distinguishing between natural variation and actual land conversion events.

\paragraph{FPCD~\cite{tundia2023fpcd}} offers an open aerial very-high-resolution dataset for farm pond change detection. The dataset contains bi-temporal imagery of agricultural landscapes with annotations for pond appearance, disappearance, and boundary changes. This dataset supports water body transition identification and quantification, enabling evaluation of agents' change detection capabilities in aquatic agricultural environments.

\paragraph{DeepWeeds~\cite{olsen2019deepweeds}} contains a multiclass weed species image dataset for deep learning applications in precision agriculture. The dataset comprises ground-level images of 9 weed species collected across diverse field conditions, with species-level classification labels. This dataset supports weed species recognition and retrieval tasks, evaluating agents' ability to identify and retrieve agronomic knowledge about specific weed species from visual evidence.

\paragraph{deepNIR~\cite{sa2022deepnir}} provides datasets for generating synthetic near-infrared images and improved fruit detection. The dataset enables fruit counting and organ detection tasks by combining RGB imagery with synthetic NIR channels, supporting quantitative estimation tasks that require precise enumeration of agricultural objects from multimodal visual inputs.

\paragraph{IP102~\cite{wu2019ip102}} is a large-scale benchmark dataset for insect pest recognition. The dataset contains over 75,000 images spanning 102 pest species, with hierarchical annotations covering pest taxonomy, morphology, and damage symptoms. This dataset supports pest identification and agricultural pest diagnosis tasks, evaluating agents' capability to recognize pests from visual evidence and retrieve relevant control measures or agronomic knowledge.

\begin{table*}[ht]
    \centering
    \caption{More details of data sources used in AgroTools.}
    \label{tab1:dataset summary}
    \scriptsize
    \renewcommand{\arraystretch}{1.15}
    \begin{tabularx}{\textwidth}{
        >{\centering\arraybackslash}m{2.2cm}
        !{\vrule width 0.5pt}
        >{\centering\arraybackslash}m{1.0cm}
        !{\vrule width 0.5pt}
        >{\raggedright\arraybackslash}X
        !{\vrule width 0.5pt}
        >{\centering\arraybackslash}m{2.2cm}
    }
        \toprule
        \textbf{Dataset} & \textbf{Year} & \textbf{Key Features} & \textbf{Dataset Link} \\
        \midrule
        WeedMap~\cite{sa2018weedmap} &
        2018 &
        Large-scale aerial multispectral imagery; semantic weed mapping annotations; precision farming scenarios& \href{https://doi.org/10.3390/rs10091423}{DOI}\\
        \midrule
        Open-Canopy~\cite{fogel2025open} &
        2025 &Very high-resolution forest monitoring data; canopy height regression; vegetation cover area segmentation &\href{https://github.com/fajwel/Open-Canopy}{GitHub} \\
        \midrule
        PlantSeg~\cite{wei2026large} &
        2026 &Large-scale in-the-wild plant disease segmentation; pixel-level lesion masks; severity estimation &\href{https://zenodo.org/records/16665770}{Zenodo}\\
        \midrule
        PApple RGB‑D Size dataset~\cite{ferrer2023simultaneous}&2023 &RGB-D images of apples; fruit detection; diameter and size estimation&\href{https://zenodo.org/records/7144328}{Zenodo}  \\
        \midrule 
       WeedSense~\cite{sarker2025weedsense} &
        2025 &Multi-task learning for weed analysis; weed segmentation; height estimation; growth stage classification &\href{https://weedsense.github.io}{Project}\\
        \midrule
        Hi-CNA~\cite{sun2024identifying}&
        2024 &Bi-temporal high-resolution remote sensing; cropland non-agriculturalization change detection &\href{http://rsidea.whu.edu.cn/Hi-CNA_dataset.htm}{Project}    \\
        \midrule    
        AgroMind~\cite{li2025can} &
        2025 &
        28,482 QA pairs and 20,850 images; four task dimensions and 13 task types; three sensor types; from nine RS datasets and one proprietary global parcel dataset &\href{https://rssysu.github.io/AgroMind/}{GitHub}        \\
        \midrule        
        CropLand-CD~\cite{liu2022cnn} &
        2022 &
        Fine-grained cropland change detection; CNN-transformer network; temporal land use analysis &\href{https://github.com/liumency/CropLand-CD}{GitHub}\\
        \midrule
        FPCD~\cite{tundia2023fpcd} &
        2023 &
        Open aerial VHR dataset; farm pond change detection; water body transition identification &
        \href{https://arxiv.org/abs/2302.14554}{arXiv}\\
        \midrule
        DeepWeeds~\cite{olsen2019deepweeds} &
        2019 &
        Multiclass weed species image dataset; 9 weed species; ground-level field images &\href{https://github.com/AlexOlsen/DeepWeeds}{GitHub}        \\
        \midrule        
        deepNIR~\cite{sa2022deepnir} &
        2022 &
        Synthetic NIR image generation; fruit detection; fruit counting; RGB and NIR fusion &\href{https://zenodo.org/records/6529979}{Zenodo}        \\
        \midrule            
        IP102~\cite{wu2019ip102} &
        2019 & Large-scale insect pest recognition; 75,000+ images; 102 pest species; hierarchical taxonomy & \href{https://github.com/xpwu95/IP102}{GitHub}\\
        \bottomrule
    \end{tabularx}
\end{table*}

\subsection{Data Integration}
The twelve datasets described above are carefully selected and integrated to create a comprehensive and problem-oriented benchmark for evaluating tool-augmented multimodal agents in agriculture. Each dataset contributes complementary strengths toward addressing the five key task families of AgroTools: recognition and retrieval, counting and measurement, segmentation interpretation, change detection, and visualization and analytics. The integration of these diverse data sources ensures that AgroTools captures a wide spectrum of agricultural scenarios, sensor modalities, and reasoning complexities, enabling rigorous evaluation of agentic capabilities in real-world agricultural settings. Specifically, the curated datasets offer the following advantages:
\paragraph{Diverse agricultural scenarios.} The datasets span multiple agricultural domains, including weed mapping, forest monitoring, plant disease diagnosis, fruit detection and measurement, pest recognition, crop change detection, and remote sensing scene understanding. This diversity ensures that AgroTools covers the major forms of agricultural problems encountered in precision farming and environmental management.
\paragraph{Multiple sensor modalities.} AgroTools integrates heterogeneous visual inputs, including standard RGB imagery, aerial multispectral data, very-high-resolution remote sensing images, RGB-D depth maps, and synthetic NIR channels. This multi-sensor coverage enables evaluation of tool-augmented agents across different imaging conditions and data types.
\paragraph{Rich annotation formats.} The source datasets provide diverse annotation types, including semantic segmentation masks, bounding boxes, species-level classification labels, pixel-level lesion masks, canopy height measurements, change detection pairs, and pest taxonomy hierarchies. These varied annotations support the construction of query-tool pairs that require different forms of visual reasoning and tool invocation.
\paragraph{Real-world relevance.} Many datasets are collected under real field conditions, capturing natural variations in lighting, occlusion, growth stages, and background complexity. This ecological validity ensures that AgroTools reflects the challenges faced by agricultural practitioners rather than simplified laboratory settings.
\paragraph{Complementary task coverage.} Each dataset contributes to specific task families within AgroTools. For example, WeedMap and DeepWeeds support recognition and retrieval tasks; PApple and deepNIR enable counting and measurement; PlantSeg supports segmentation interpretation; Hi-CNA, CropLand-CD, and FPCD facilitate change detection; and AgroMind provides multi-dimensional coverage across multiple task types.
\paragraph{Scalability and diversity of scale.} The datasets range from small-scale specialized collections (e.g., PApple with focused fruit measurement) to large-scale comprehensive benchmarks (e.g., IP102 with over 75,000 images, AgroMind with 20,850 images), enabling evaluation across different data regimes and generalization requirements.

\subsection{Data Pre-processing Details}
To accommodate the substantial heterogeneity across the 12 source datasets, we adopted a unified pre-processing pipeline to standardize imagery, annotations, and sample organization prior to benchmark construction. Whenever official train/validation/test splits were available, we preserved them. For ultra-high-resolution imagery such as WeedMap, we followed the original cropping protocol and naming convention to derive patch-level samples. Across datasets, raw inputs were converted into task-consistent formats by aligning each image with its corresponding annotation, harmonizing label representations, filtering invalid or incomplete records, and normalizing auxiliary metadata. For sources involving additional modalities or structured supervision, such as depth maps, bi-temporal image pairs, and polygon- or box-level annotations, we further transformed the original labels into a common representation suitable for downstream evaluation while preserving task-specific semantics. After pre-processing, all retained samples were organized into a unified benchmark schema, in which images, annotations, and associated metadata are stored in a consistent JSON-compatible format, thereby enabling reliable evaluation across segmentation, change detection, counting, phenotyping, and classification tasks.

\paragraph{WeedMap Dataset.} For WeedMap, we first read the raw multi-channel orthomosaic imagery together with the corresponding ground-truth maps at the scene level, and aligned each image tile with its annotation. Because the original imagery is ultra-high resolution, we followed the official cropping strategy and naming convention to partition each scene into fixed-size patches. We further used the reflectance maps to identify invalid regions and marked no-data pixels as ignore labels. After removing samples with incomplete image, label, or metadata fields, we retained patch-level metadata such as scene ID, crop position, and split assignment, resulting in a standardized patch dataset for downstream weed segmentation and reasoning tasks. 

\paragraph{Open-Canopy Dataset.} For Open-Canopy, we used the original spatial geometries and official split information to associate each sample with its corresponding SPOT imagery and LiDAR measurements. We then cropped local regions using the polygon bounds provided in the source annotations and resampled both imagery and height data to a unified input size. For canopy height estimation, LiDAR values were converted from decimeters to meters and invalid land-cover classes were filtered out using the accompanying classification layer; for canopy delineation, pixels above a fixed height threshold were converted into binary canopy masks. Finally, we retained only fully aligned test samples and exported them as uniformly named GeoTIFF patches for the canopy-related benchmark tasks.

\paragraph{PlantSeg Dataset.} For PlantSeg, we directly adopted the official train/validation/test partition and paired each image with its annotation by filename stem. Since the annotations primarily describe diseased regions, we converted all non-zero mask values into a binary lesion foreground, thereby standardizing the source masks into a unified lesion segmentation format. We then resized images and masks to a fixed resolution and removed incomplete pairs. For benchmark construction, we sampled from the standardized test split and anonymized the exported filenames while preserving a mapping to the original identifiers for traceability.

\paragraph{PApple RGB-D Size Dataset.} For the PApple RGB-D Size dataset, we jointly loaded aligned RGB images, depth maps, camera focal lengths, and ground-truth apple diameters under the official split structure. Instance annotations were parsed from VIA-style polygon files, and multiple regions sharing the same apple\_ID were merged while invalid IDs were removed. We then converted polygon annotations into instance masks and rectangular regions suitable for downstream processing, and matched each sample to its corresponding diameter annotation using filename-derived metadata. Only samples with valid RGB, depth, focal length, and instance annotations were retained, yielding a standardized RGB-D benchmark subset for apple size estimation and ROI-based reasoning.

\paragraph{WeedSense Dataset.} For WeedSense, we reorganized the raw data into train, validation, and test splits and aligned each image with its segmentation mask and tabular metadata entry. When the original mask was provided as a binary foreground map, we remapped foreground pixels to species-specific semantic labels using the species field in the accompanying CSV file, thereby unifying heterogeneous annotations into a common label space. We also preserved plant height measurements and normalized the original week labels into a contiguous growth-stage index. After standardization, we further grouped samples by species and temporal order to support multi-image phenotyping and curve-based reasoning tasks in the benchmark.

\paragraph{Hi-CNA Dataset.} For Hi-CNA, we paired the two temporal images for each sample with the corresponding change annotation, and retained per-time-step semantic labels when available. Because the source imagery is stored as multi-band TIFF files, we applied consistent channel reading and intensity scaling to standardize the dynamic range across samples. We then binarized the change labels and resized both images and annotations to a fixed resolution. Samples with missing temporal pairs or incomplete annotations were discarded, producing a unified paired-image change detection subset for benchmark evaluation.

\paragraph{AgroMind.} For AgroMind, we used the aerial palm-tree images annotated with red bounding boxes to construct the benchmark subset. During preprocessing, we retained the target-tree localization cues encoded by the red boxes, standardized the file naming and sample indexing, and filtered out images with ambiguous targets, incomplete annotations, or insufficient visual quality. The resulting subset was used for tree counting and target-specific spatial reasoning tasks.

\paragraph{CropLand-CD Dataset.} For CropLand-CD, we followed the official train/validation/test split and matched each time1 image, time2 image, and change label by filename. We converted all non-zero label values into a binary change mask to eliminate inconsistencies in raw label encoding across samples. Both temporal images and labels were then resized to a common spatial resolution, and incomplete triplets were removed. This yielded a standardized bi-temporal change detection subset for cropland change analysis in the benchmark.

\paragraph{FPCD Dataset.} For FPCD, we used the official train and test file lists to match the two temporal images with their associated change masks at the sample level. Because the original annotations contain multi-class change information, we preserved the original semantic labels where appropriate and additionally derived binary change masks when a unified change-detection interface was required. We further split a validation subset from the training data using a fixed random seed to maintain a consistent experimental protocol, and resized all images and labels to a common resolution. Only complete image-pair and label triplets were retained for the final benchmark subset.

\paragraph{DeepWeeds Dataset.} For DeepWeeds, we used the official class annotation file together with the provided five-fold split definitions, while preserving the original nine-way label space, including the negative class. Since the dataset is natively formulated as an image-level classification task, preprocessing mainly consisted of label normalization, official split preservation, and removal of incomplete or invalid records rather than pixel-level annotation conversion. We then selected representative test images from the standardized data and reorganized them into the single-image and multi-image formats required by the benchmark, while keeping the original class semantics unchanged.

\paragraph{deepNIR Dataset.} For deepNIR, we unified all 11 fruit subsets under a common preparation pipeline and preserved the original train/validation/test partition. The source annotations are provided in YOLO format, so we converted normalized bounding-box coordinates into absolute image-space rectangles while retaining the subset name as the category identity. We then filtered out samples with missing labels or visually ambiguous targets and standardized all subsets into a common object-counting data format. From this unified pool, we selected representative images for the fruit counting tasks in the benchmark.

\paragraph{IP102 Dataset.} During preprocessing, we standardized class names and image formats, converted uncommon formats into a common representation, and generated leakage-aware train/validation/test manifests with content-hash checking to prevent near-duplicate images from crossing splits. We also enforced class-level data sufficiency constraints to ensure stable downstream learning. Finally, we selected well-defined, crop-relevant pest categories from the standardized pool and exported them into the pest/crop classification portion of the benchmark while preserving image-to-label mappings for traceability.

\subsection{Question and Answer Examples}
\begin{figure}
    \centering
    \includegraphics[width=1\linewidth]{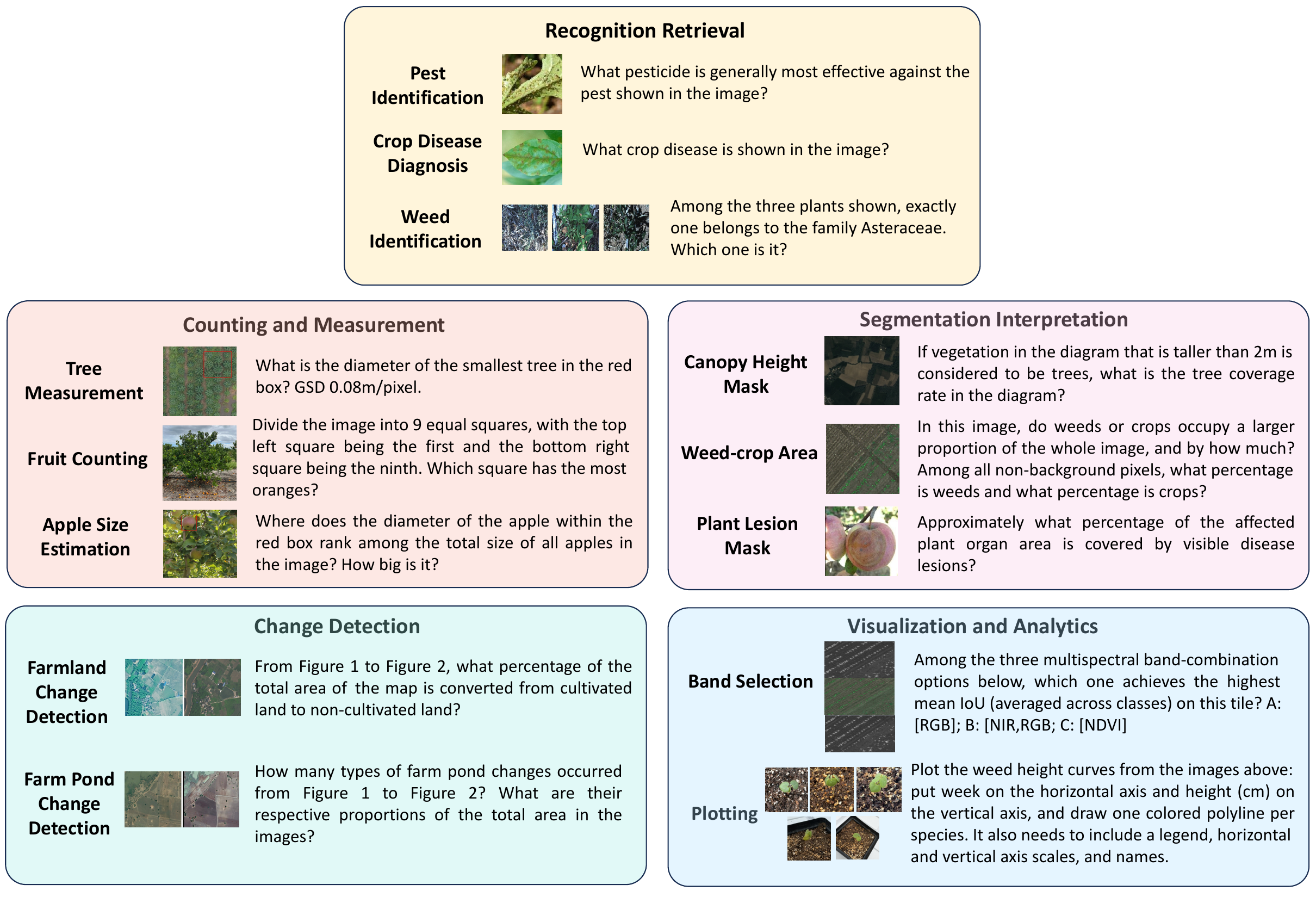}
    \caption{Some representative examples from the five task categories.}
    \label{fig6}
\end{figure}

We showcase representative examples spanning all five core task categories of AgroTools in Fig.~\ref{fig6}, each designed to evaluate distinct multimodal capabilities required for real-world agricultural problem-solving.

The Recognition Retrieval category tests foundational visual identification and domain knowledge lookup, including pest identification (e.g., recommending appropriate pesticides based on pest morphology), crop disease diagnosis, and weed species classification, which require matching visual features with specialized agricultural knowledge. Next, Visualization and Analytics assesses higher-order analytical skills, such as selecting optimal multispectral band combinations for segmentation tasks and generating structured plots (e.g., weed height curves with proper labeling and scaling), mirroring real-world agricultural data analysis workflows. The Counting and Measurement category focuses on quantitative spatial reasoning, requiring models to compute tree diameters (using provided ground sampling distance), count fruits in defined grid cells, and estimate the size and rank of individual objects relative to the entire population. For Change Detection, tasks involve comparing multi-temporal imagery to quantify land use conversions (e.g., cultivated to non-cultivated land) and classify farm pond changes, testing the model’s ability to track spatial and temporal variations in agricultural landscapes. Finally, Segmentation Interpretation evaluates semantic understanding of segmentation outputs, such as calculating tree coverage rates from height masks, comparing weed-crop area proportions, and estimating the percentage of plant tissue affected by lesions, requiring both visual parsing and arithmetic reasoning. Across all categories, the tasks range from basic visual recognition to complex multi-step reasoning and tool use, collectively challenging models to integrate domain knowledge, spatial analysis, and structured output generation—key capabilities for robust agricultural multimodal agents.



\section{Tool Definition}

\subsection{Tool Overview}
Table~\ref{tab6-tool-overview} provides a compact overview of the tools used in our framework. The tools span computation, visualization, retrieval, visual understanding, image processing, and agricultural decision support, covering the main capabilities needed for multi-step agricultural reasoning. Calculator and Plot are implemented with a Python execution environment and are deterministic under the same inputs; ExtractMarkedRedBox is also deterministic because it follows a rule-based red-box extraction procedure. In contrast, AgriculturalWebSearch and ImageRegionUnderstanding depend on external APIs, and the remaining perception and decision tools are supported by different models, making their outputs potentially non-deterministic. By exposing each tool's function, input-output format, determinism, and backend, the table clarifies how complex agricultural questions are grounded into executable tool-use steps.

\subsection{Tool Details.}
\label{appendix:tool details}
In this section, all tool names are presented using their original identifiers as defined in our codebase. For ease of cross-reference, the corresponding tool names shown in Fig.~\ref{fig2} are explicitly indicated in parentheses in the text.

\paragraph{Calculator} (Calculator) is a deterministic symbolic-numeric utility implemented as a restricted Python-expression evaluator. It supports standard arithmetic, elementary mathematical operators, and a limited set of safe built-in functions (e.g., max, min, round, and functions from the math namespace), while disallowing arbitrary package imports. In our framework, this tool is primarily used to perform intermediate quantitative reasoning, such as area computation, ratio estimation, unit conversion, and simple descriptive statistics. Since it is purely rule-based and executes under a fixed timeout, it does not involve model training.

\paragraph{AgriculturalWebSearch} (WebSearch) is an external retrieval tool designed for agriculture-oriented factual queries. It wraps the Tavily search API and prepends a domain-specific prompt so that retrieval is biased toward agronomic evidence, including pest taxonomy, host plants, disease etiology, geographic distribution, climate suitability, and control recommendations. The tool returns a concise textual answer together with source links, thereby providing lightweight evidence grounding for knowledge-intensive questions. As it relies on live web retrieval rather than local model fitting, no task-specific training is required. We avoid direct use of Google Search, since it often yields irrelevant content and garbled files that disturb model reasoning. We manually validate this tool to guarantee trustworthy outputs.

\paragraph{Plot} (Plot) is a deterministic Python-based visualization tool that executes short plotting scripts and returns the resulting figure as an image. The user-provided code is required to define a solution() function that directly returns a matplotlib figure object, allowing the framework to convert structured numerical outputs into publication-style charts. In practice, the tool is used for grouped bar charts, line plots, and other visual summaries required by the benchmark. It is not learned and therefore does not involve any training procedure.

\paragraph{ObjectCounting} (ObjectCount) is a unified counting module with category-aware backend dispatch. For tree-like targets, such as trees, pines, palms, or canopy objects, it uses an open-vocabulary detector, with Grounding-DINO Base as the default backend and YOLO-World v2 as an optional alternative; this branch is not fine-tuned locally and instead relies on open-vocabulary detection combined with ROI-aware filtering, optional upscaling, tiled inference, and non-maximum suppression. For fruit categories, the tool switches to task-specific YOLOv8m detectors fine-tuned on the prepared CITDET orange dataset and the deepNIR 11-fruit dataset, respectively. Both fruit detectors were trained under the configuration in ObjectCounting\_finetune, using a 40-epoch schedule, batch size 3, learning rate 0.005, weight decay 0.0005, and model selection based on AP50. This hybrid design allows the framework to retain open-vocabulary flexibility for trees while using stronger specialized detectors for fruit-counting tasks.

\paragraph{ImageRegionUnderstanding} (ImageDescripe) is a region-level visual interpretation tool powered by GPT-5.4 through an OpenAI-compatible Responses endpoint. Given an image and a bounding box (optional), it crops the specified region and submits only that crop to the vision-language model, thereby focusing the response on local content rather than global scene context. The tool is used to describe spatial layout, object attributes, land-cover patterns, and other semantically grounded properties inside a marked region. As an API-based vision component, it is not trained within our framework.

\paragraph{PestCropIdentification} (PestIdentify) is an image-level classifier for agricultural pests and crop diseases. The deployed model is a ConvNeXtV2-Base classifier (\texttt{convnextv2\_base.fcmae\_ft\_in22k\_in1k}) with an input resolution of 384, trained on the unified pest-and-crop classification pool constructed from multiple public agricultural datasets. The released checkpoint used by the tool was trained under a 40-epoch schedule with batch size 48, learning rate $3 \times 10^{-4}$, weight decay 0.05, weighted sampling, RandAugment-based augmentation, and label smoothing; the best model was saved at epoch 18 according to validation macro-F1. At inference time, the tool returns the top-1 predicted label, providing a compact interface for pest and disease recognition questions.

\paragraph{ExtractMarkedRedBox} (ExtractRedBox) is a deterministic rule-based utility for recovering manually drawn red rectangular markers from benchmark images. Rather than relying on learned detection, it combines strict and relaxed red-color masking, connected-component extraction, component merging, geometric refinement, and fallback heuristics to robustly identify annotated red boxes under mild variations in line thickness and color saturation. This tool is particularly important for questions in which a region of interest is highlighted by a human-drawn red rectangle. Because the target signal is a synthetic annotation rather than a natural object category, a rule-based implementation is both simpler and more stable than a learned detector.

\paragraph{AppleSizeEstimate} (AppleSizeEstimate) is an RGB-D instance-level apple sizing tool built on top of a four-channel Mask R-CNN with ResNet50-FPN. The implementation adapts the standard torchvision \texttt{maskrcnn\_resnet50\_fpn} architecture so that the first convolution accepts four channels, allowing aligned RGB and depth inputs to be processed jointly; predicted masks are then combined with the depth map and camera focal length to estimate real-world apple diameter. The model was trained on the \texttt{PApple RGB-D Size} dataset using a 40-epoch schedule, batch size 2, learning rate 0.005, momentum 0.9, weight decay $10^{-4}$, and Mask R-CNN resize settings of \texttt{min\_size=800} and \texttt{max\_size=1333}; model selection follows the validation detection metric with geometric error used as a secondary criterion. The tool outputs per-instance diameter estimates, optional bounding boxes, and an annotated visualization path.

\paragraph{PlantSegLesionAnalysis} (LesionSegment) is a two-stage lesion interpretation tool that combines disease-region segmentation with organ-aware mask refinement. The lesion predictor is a \texttt{SegFormer} model with a \texttt{MiT-B4} encoder, trained on the standardized \texttt{PlantSeg} lesion dataset for 40 epochs with batch size 4, learning rate $3 \times 10^{-4}$, weight decay $10^{-4}$, and input resolution $512 \times 512$; the released best checkpoint corresponds to epoch 31. To obtain a biologically more coherent organ mask, the tool further applies a lesion-guided \texttt{SAM2.1 Hiera-Large} model, using the predicted lesion regions to select relevant SAM masks. We carefully checked all the images related to the problem to ensure that the organ masks segmented by the SAM2.1 model were correct. Images with incorrect segmentation were discarded.

\paragraph{ChangeDetection} (Non-farmland Detect, Farm pond types Detect) is a unified interface over three dataset-specific bi-temporal change-detection models. For \texttt{Hi-CNA}, it uses a four-band \texttt{dual\_resnet\_fpn} model with a \texttt{ResNet34} encoder, trained on $512 \times 512$ inputs under a 40-epoch schedule with batch size 4. For \texttt{CropLand-CD} (\texttt{CLCD}), it uses a \texttt{ResNet34} baseline change detector trained on $512 \times 512$ RGB image pairs for 40 epochs with batch size 4, learning rate $3 \times 10^{-4}$, and weight decay $10^{-4}$, with the released best checkpoint saved at epoch 31. For \texttt{FPCD}, it employs a five-class \texttt{dual\_resnet\_fpn} model with a \texttt{ResNet34} encoder, trained for 40 epochs with batch size 4, learning rate $3 \times 10^{-4}$, weight decay $10^{-4}$, and a validation split ratio of 0.15. This design allows the same tool interface to support binary cropland change detection, four-band high-resolution change analysis, and multi-class farm-pond transition recognition.

\paragraph{CanopyHeightEstimate} (CanopyHeightEstimate) is a per-pixel canopy-height regression tool that predicts vegetation height from a single multi-band GeoTIFF canopy patch. The deployed model is a \texttt{custom\_resnet34\_unet}, i.e., a U-Net-style regression network with a \texttt{ResNet34} encoder, trained on the \texttt{Open-Canopy} height-estimation dataset. The checkpoint used by the tool was trained for 30 epochs with batch size 16, learning rate $3 \times 10^{-4}$, weight decay $10^{-4}$, four input channels, and $256 \times 256$ resized patches; the released best checkpoint was saved at epoch 28 and uses 2.0 m as the default tree-cover threshold. In addition to height statistics, the tool can export an RGB preview, a height visualization, a binary canopy mask, and the predicted floating-point height array.

\paragraph{WeedSpeciesClassification} (WeedClassify) is a weed-image classifier trained on \texttt{DeepWeeds}. Like the pest-and-disease classifier, it uses \texttt{ConvNeXtV2-Base} at 384 resolution, but here the label space is the original nine-way DeepWeeds taxonomy, including eight weed categories and one negative class. The deployed checkpoint was trained with a 40-epoch schedule, batch size 32, learning rate $3 \times 10^{-4}$, weight decay 0.05, weighted sampling, label smoothing, and standard augmentation, and the released model corresponds to fold 0 with the best checkpoint saved at epoch 32. The tool supports either single-image or multi-image input and returns the predicted class label(s) in the input order.

\paragraph{WeedPhenotypingAnalysis} (WeedHeightAnalyze) is the benchmark’s multi-task weed phenotyping tool, built on the \texttt{WeedSense} model. It jointly predicts semantic weed class, plant height, and growth-stage week from RGB imagery, and subsequently summarizes the semantic output as the dominant non-background class for downstream reasoning. The released model was trained on the \texttt{WeedSense} dataset with $512 \times 512$ inputs for 40 epochs, batch size 8, AdamW optimization with learning rate $2 \times 10^{-4}$ and weight decay $10^{-4}$, and a warmup-cosine learning-rate schedule; the checkpoint packaged with the tool was saved at epoch 34. The multitask loss uses equal weights for segmentation, height regression, and week classification, together with auxiliary supervision, making this tool well suited to temporal phenotyping and growth-curve analysis.

\paragraph{WeedSegmentationAnalysis} (WeedSegment) is a model-library-style segmentation tool built for \texttt{WeedMap} rather than a single fixed model. Internally, it maintains a suite of \texttt{SegNet} models trained on different spectral band combinations from the WeedMap RedEdge imagery, and at inference time it resolves the requested band combination to the best-performing checkpoint according to a precomputed summary of test performance. Across this model suite, training follows a largely shared protocol: RedEdge tiles of size $360 \times 480$, training scenes \texttt{000/001/002/004}, test scene \texttt{003}, a 40,000-iteration schedule, SGD optimization with learning rate $10^{-3}$, momentum 0.9, weight decay $5 \times 10^{-4}$, Poly decay, horizontal flipping, and class-balanced training; the batch size varies slightly across band combinations, typically between 5 and 6. The tool can therefore return both prediction-only pixel statistics and ground-truth-based segmentation metrics while preserving the spectral specificity of the original WeedMap experiments.

\begin{table}[t]
\centering
\small
\setlength{\tabcolsep}{5pt}
\renewcommand{\arraystretch}{1.08}
\caption{
Overview of the tools used in our framework.
}
\label{tab6-tool-overview}
\begin{adjustbox}{max width=\textwidth,max totalheight=0.82\textheight,keepaspectratio}
\begin{tabular}{ccccc}
\toprule
\multicolumn{1}{c|}{\textbf{Tool}} &
\multicolumn{1}{c|}{\textbf{Input}} &
\multicolumn{1}{c|}{\textbf{Output}} &
\multicolumn{1}{c|}{\textbf{Det.}} &
\textbf{Backend} \\
\midrule

\multicolumn{1}{c|}{\textsc{Calculator}} &
\multicolumn{1}{c|}{Text / Expr.} &
\multicolumn{1}{c|}{Text / Number} &
\multicolumn{1}{c|}{Yes} &
Py \\

\midrule
\multicolumn{1}{c|}{\textsc{AgriculturalWebSearch}} &
\multicolumn{1}{c|}{Text Query} &
\multicolumn{1}{c|}{Text / Evidence} &
\multicolumn{1}{c|}{No} &
API-S \\

\midrule
\multicolumn{1}{c|}{\textsc{Plot}} &
\multicolumn{1}{c|}{Code / Data} &
\multicolumn{1}{c|}{Image} &
\multicolumn{1}{c|}{Yes} &
Py \\

\midrule
\multicolumn{1}{c|}{\textsc{ObjectCounting}} &
\multicolumn{1}{c|}{Image / ROI / Text} &
\multicolumn{1}{c|}{JSON / Image Path} &
\multicolumn{1}{c|}{No} &
Model \\

\midrule
\multicolumn{1}{c|}{\textsc{ImageRegionUnderstanding}} &
\multicolumn{1}{c|}{Image / BBox / Text} &
\multicolumn{1}{c|}{Text} &
\multicolumn{1}{c|}{No} &
API-V \\

\midrule
\multicolumn{1}{c|}{\textsc{PestCropIdentification}} &
\multicolumn{1}{c|}{Image} &
\multicolumn{1}{c|}{Label} &
\multicolumn{1}{c|}{No} &
Model \\

\midrule
\multicolumn{1}{c|}{\textsc{ExtractMarkedRedBox}} &
\multicolumn{1}{c|}{Image} &
\multicolumn{1}{c|}{JSON / BBox} &
\multicolumn{1}{c|}{Yes} &
Rule \\

\midrule
\multicolumn{1}{c|}{\textsc{AppleSizeEstimate}} &
\multicolumn{1}{c|}{RGB-D / ROI} &
\multicolumn{1}{c|}{JSON / Image Path} &
\multicolumn{1}{c|}{No} &
Model \\

\midrule
\multicolumn{1}{c|}{\textsc{PlantSegLesionAnalysis}} &
\multicolumn{1}{c|}{Image} &
\multicolumn{1}{c|}{JSON / Image Path} &
\multicolumn{1}{c|}{No} &
Model \\

\midrule
\multicolumn{1}{c|}{\textsc{ChangeDetection}} &
\multicolumn{1}{c|}{Image Pair / ROI} &
\multicolumn{1}{c|}{JSON / Image Path} &
\multicolumn{1}{c|}{No} &
Model \\

\midrule
\multicolumn{1}{c|}{\textsc{CanopyHeightEstimate}} &
\multicolumn{1}{c|}{GeoTIFF / Text} &
\multicolumn{1}{c|}{JSON / Image Path} &
\multicolumn{1}{c|}{No} &
Model \\

\midrule
\multicolumn{1}{c|}{\textsc{WeedSpeciesClassification}} &
\multicolumn{1}{c|}{Image(s)} &
\multicolumn{1}{c|}{Label(s)} &
\multicolumn{1}{c|}{No} &
Model \\

\midrule
\multicolumn{1}{c|}{\textsc{WeedPhenotypingAnalysis}} &
\multicolumn{1}{c|}{Image(s)} &
\multicolumn{1}{c|}{JSON} &
\multicolumn{1}{c|}{No} &
Model \\

\midrule
\multicolumn{1}{c|}{\textsc{WeedSegmentationAnalysis}} &
\multicolumn{1}{c|}{Image(s) / Text} &
\multicolumn{1}{c|}{JSON / Image Path} &
\multicolumn{1}{c|}{No} &
Model \\

\bottomrule
\end{tabular}
\end{adjustbox}

\vspace{2pt}
\begin{minipage}{\textwidth}
\raggedright
\footnotesize
\emph{Note.}
\textbf{Tool} denotes the tool name;
\textbf{Input} and \textbf{Output} summarize high-level formats rather than detailed parameters.
\textbf{Det.} indicates whether the same input is expected to produce the same output.
\textbf{Backend}: Py = Python execution environment; API-S = external search API; API-V = external vision API; Rule = rule-based processing; Model = model-supported inference.
\end{minipage}

\end{table}

\section{Experiment Details}

\subsection{Details of Evaluation Metrics}
\label{appendix:evaluation metrics}
We evaluate AgroTools from two complementary perspectives. Process-level metrics are used in the step-by-step setting to measure whether a model follows the reference interaction trajectory. Answer-level metrics are used in the no-tool and end-to-end settings to measure the quality of the final answer. All reported scores except EAR are multiplied by 100 in the tables for readability.

\paragraph{Process-level metrics.}
In step-by-step evaluation, each model is conditioned on the preceding gold interaction history and is asked to predict the next assistant action. Each gold action is either a tool call or a final-answer response. We report five process-level metrics.

\textbf{Step-Type Accuracy (STAcc).}
STAcc measures whether the model predicts the correct next action type, namely whether it should continue with a tool call or stop and produce an answer:
\begin{equation}
    \mathrm{STAcc}
=
\frac{1}{N_{\mathrm{step}}}
\sum_{i,t}
\mathbb{I}\!\left[\hat{z}_{it}=z_{it}\right],
\end{equation}
where \(z_{it}\) and \(\hat{z}_{it}\) denote the gold and predicted action types at step \(t\) of sample \(i\), and \(N_{\mathrm{step}}\) is the total number of evaluated steps.

\textbf{Tool Accuracy (ToolAcc).}
ToolAcc is computed only on steps where the gold action is a tool call. A prediction is counted as correct only when the model both decides to call a tool and selects the same tool as the reference:

\begin{equation}
    \mathrm{ToolAcc}
=
\frac{1}{N_{\mathrm{tool}}}
\sum_{(i,t)\in \mathcal{C}}
\mathbb{I}\!\left[
\hat{z}_{it}=\mathrm{tool}
\wedge
\hat{a}_{it}=a_{it}
\right],
\end{equation}
%
where \(\mathcal{C}\) denotes the set of gold tool-call steps, \(a_{it}\) and \(\hat{a}_{it}\) are the gold and predicted tool names, and \(N_{\mathrm{tool}}=|\mathcal{C}|\).

\textbf{Argument Accuracy (ArgAcc).}
ArgAcc further requires the predicted arguments to match the reference arguments. It therefore reflects whether a model can not only select the correct tool, but also instantiate it with valid and task-consistent parameters:
\begin{equation}
    \mathrm{ArgAcc}
=
\frac{1}{N_{\mathrm{tool}}}
\sum_{(i,t)\in \mathcal{C}}
\mathbb{I}\!\left[
\hat{z}_{it}=\mathrm{tool}
\wedge
\hat{a}_{it}=a_{it}
\wedge
\mathrm{MatchArgs}(\hat{\mathbf{u}}_{it},\mathbf{u}_{it})
\right].
\end{equation}

Here \(\mathbf{u}_{it}\) and \(\hat{\mathbf{u}}_{it}\) denote the gold and predicted argument objects. In practice, arguments are compared after standard normalization. For tools whose arguments naturally allow multiple surface forms, such as search queries and plotting commands, we use relaxed normalized string matching.

\textbf{Early Answer Rate (EAR).}
EAR measures how often the model produces a final answer before the required tool-use procedure is completed:
\begin{equation}
    \mathrm{EAR}
=
\frac{1}{N_{\mathrm{tool}}}
\sum_{(i,t)\in \mathcal{C}}
\mathbb{I}\!\left[
\hat{z}_{it}=\mathrm{answer}
\right].
\end{equation}

Lower EAR is better. We report EAR as a diagnostic metric for premature stopping.

\textbf{Summary Accuracy (SummAcc).}
SummAcc evaluates whether the model can produce the correct final answer when conditioned on the preceding gold interaction history. It is computed on gold answer steps. If the model predicts a tool call instead of an answer, the score for that step is zero; otherwise, the predicted answer is compared with the reference answer using normalized textual similarity, with additional penalties for numeric inconsistency and missing key content:
\begin{equation}
    \mathrm{SummAcc}
=
\frac{1}{N_{\mathrm{ans}}}
\sum_{(i,t)\in \mathcal{A}}
s^{\mathrm{sum}}_{it},
\qquad
s^{\mathrm{sum}}_{it}\in[0,1],
\end{equation}

where \(\mathcal{A}\) is the set of gold answer steps. Although reported as an accuracy-style metric for consistency with the main table, SummAcc is a normalized summary score in \([0,1]\).

\paragraph{Answer-level slot scoring.}
For no-tool and end-to-end evaluation, we score the final answer through a slot-based evaluator. Each sample is decomposed into one or more answer slots, and each slot receives a score in \([0,1]\). Closed-form slots are scored deterministically, while open-text slots are scored by an LLM judge. Details of the LLM judge are provided in Appendix~\ref{appendix:llm_evaluator}.

Closed-form slots include exact text labels, counts, numerical values, lists, and key-value maps. Text slots are evaluated by normalized exact or alias matching. Count and numerical slots are evaluated with exact integer matching or a fixed numerical tolerance. List-valued and map-valued slots receive partial credit based on matched elements. Open-text slots are evaluated by a strict LLM judge according to factual coverage, specificity, and contradiction against the reference key points. The raw judge score is used directly in the aggregate metrics.

\paragraph{Final Answer Score (FAS).}
Final Answer Score is the average sample-level final-answer score. For ordinary structured or text-answer samples, the sample-level score is the mean of its slot scores:
\begin{equation}
    \mathrm{FAS}_i
=
\frac{1}{|\mathcal{S}_i|}
\sum_{j\in\mathcal{S}_i}
s_{ij},
\end{equation}

where \(\mathcal{S}_i\) is the set of scored answer slots for sample \(i\), and \(s_{ij}\in[0,1]\) is the score of slot \(j\). The dataset-level FAS is then
\begin{equation}
    \mathrm{FAS}
=
\frac{1}{|\mathcal{D}|}
\sum_{i\in\mathcal{D}}
\mathrm{FAS}_i.
\end{equation}


The evaluation set \(\mathcal{D}\) depends on the reported result. In Table~\ref{tab3:main_results}, no-tool and end-to-end results are reported on the same shared subset for fair comparison, excluding samples that require raster-format image processing or plot generation. Therefore, the Table~\ref{tab3:main_results} FAS does not include plot-generation samples. In task-level analysis over the full dataset, such as Fig.~\ref{fig5}, plot-generation samples are included by replacing \(\mathrm{FAS}_i\) with the dedicated plot score defined below.

\paragraph{Closed-Slot Score (CSS).}
Closed-Slot Score averages scores over all closed-form slots:
\begin{equation}
    \mathrm{CSS}
=
\frac{1}{N_{\mathrm{closed}}}
\sum_i
\sum_{j\in\mathcal{S}^{\mathrm{closed}}_i}
s_{ij}.
\end{equation}

CSS is not a strict exact-match accuracy, because list-valued and structured numerical slots may receive partial credit.

\paragraph{Open-Text Score (OTS).}
Open-Text Score averages scores over all open-text slots:
\begin{equation}
    \mathrm{OTS}
=
\frac{1}{N_{\mathrm{open}}}
\sum_i
\sum_{j\in\mathcal{S}^{\mathrm{open}}_i}
s^{\mathrm{open}}_{ij},
\end{equation}

where \(s^{\mathrm{open}}_{ij}\in[0,1]\) is the raw LLM-judge score.

\paragraph{Plot evaluation.}
Plot-generation samples are evaluated by a dedicated plot evaluator. For each plot sample, we compute a visual score \(v_i\in[0,1]\) that measures whether the generated figure matches the reference plot, and a parameter score \(p_i\in[0,1]\) that measures whether the underlying plotting command and data are consistent with the reference. The final plot score is
\begin{equation}
    s^{\mathrm{plot}}_i
=
\frac{v_i+p_i}{2}.
\end{equation}

If no valid plot is generated, the plot score is set to zero. The aggregate Plot score is
\begin{equation}
    \mathrm{Plot}
=
\frac{1}{|\mathcal{D}_{\mathrm{plot}}|}
\sum_{i\in\mathcal{D}_{\mathrm{plot}}}
s^{\mathrm{plot}}_i.
\end{equation}

When reporting full-dataset task-level FAS, plot-generation samples use \(s^{\mathrm{plot}}_i\) as their sample-level final-answer score.

\subsection{Evaluation Models}
\begin{table}[hptb]
    \centering
    \caption{Models grouped by family.}
    \label{tab:model_comparison_grouped}
    \scriptsize
    \begin{tabularx}{\linewidth}{
    >{\centering\arraybackslash}m{2cm}
    !{\vrule width 0.5pt}
    >{\centering\arraybackslash}m{3.2cm}
    !{\vrule width 0.5pt}
    >{\centering\arraybackslash}m{1.0cm}
    !{\vrule width 0.5pt}
    >{\centering\arraybackslash}X
}
        \toprule
        \textbf{Model Family} & \textbf{Model Name} & \textbf{Year} & \textbf{Parameters} \\
        
        \specialrule{\lightrulewidth}{0pt}{0pt}
        \rowcolor[HTML]{FCE4D6}
        \multicolumn{4}{l}{
          \rule{0pt}{3.0ex}\textbf{Proprietary, API}\rule[-1.5ex]{0pt}{3.0ex}
        } \\
        \specialrule{\lightrulewidth}{0pt}{0.6ex}

        GPT
          & GPT-5.4 & 2026 & N/A \\
        \midrule

        Claude
          & Claude-Sonnet-4.6 & 2026 & N/A \\
        \midrule

        Gemini
          & Gemini-2.5-Pro & 2025 & N/A \\
        \midrule

        Doubao
          & Doubao-Seed-2.0-pro & 2026 & N/A \\

        \specialrule{\lightrulewidth}{0pt}{0pt}
        \rowcolor[HTML]{FFF2CC}
        \multicolumn{4}{l}{
          \rule{0pt}{3.0ex}\textbf{Open-source}\rule[-1.5ex]{0pt}{3.0ex}
        } \\
        \specialrule{\lightrulewidth}{0pt}{0.6ex}

        \multirow[c]{3}{*}{InternVL}
          & InternVL3-8B & 2025 & 8B \\
          & InternVL3.5-8B & 2025 & 8B \\
          & InternVL3.5-14B & 2025 & 14B \\
        \midrule

        \multirow[c]{3}{*}{Qwen-VL}
          & Qwen3.5-9B & 2026 & 9B \\
          & Qwen3.5-35B-A3B & 2026 & 35B \\
          & Qwen3-VL-8B-Instruct & 2025 & 8B \\
        \midrule

        LLaVA-NeXT
          & LLaVA-NeXT-13B & 2024 & 13B \\
        \midrule

        DeepSeek
          & Janus-Pro-7B & 2025 & 7B \\
        \midrule

        GLM-V
          & GLM-4.6V & 2026 & 106B \\
        \bottomrule
    \end{tabularx}
\end{table}

As shown in Tab.~\ref{tab:model_comparison_grouped}, we evaluate a diverse set of proprietary and open-source vision-language models covering a range of model families and parameter scales. For families with multiple versions or model sizes, we include representative variants to examine the effect of model scaling and model iteration on multimodal reasoning performance.

\paragraph{Proprietary Models.}
We consider four major proprietary multimodal systems accessed via their official APIs: GPT-5.4, Claude-Sonnet-4.6, Gemini-2.5-Pro, and Doubao-Seed-2.0-pro. These models support joint image-text input and long-form natural-language generation, and they serve as strong commercial baselines for evaluating zero-shot multimodal reasoning ability.

\textbf{GPT} is represented by GPT-5.4, a closed-source multimodal model designed for general-purpose reasoning and instruction following.

\textbf{Claude} is represented by Claude-Sonnet-4.6, which emphasizes robust language understanding, visual comprehension, and stable multimodal interaction.

\textbf{Gemini} is represented by Gemini-2.5-Pro, a proprietary multimodal model optimized for image-text understanding and complex reasoning.

\textbf{Doubao} is represented by Doubao-Seed-2.0-pro, a commercial multimodal system designed for general visual understanding and question answering.

\paragraph{Open-source Models.}
We further evaluate several open-source multimodal model families, including InternVL, Qwen-VL, LLaVA-NeXT, DeepSeek, and GLM-V. These models differ in architecture, scale, and training strategy, providing a broad view of current open-source multimodal systems.

\textbf{InternVL} is a family of open-source vision-language models that tightly couples visual perception and language generation in a unified framework. We evaluate three variants, InternVL3-8B, InternVL3.5-8B, and InternVL3.5-14B, to study both version-level improvements and the effect of scaling within the same family.

\textbf{Qwen-VL} is an instruction-tuned multimodal model family for image-text understanding, visual dialogue, and reasoning. We include Qwen3.5-9B, Qwen3.5-35B-A3B, and Qwen3-VL-8B-Instruct as representative variants with different model capacities.

\textbf{LLaVA-NeXT} extends the LLaVA framework with stronger language backbones and improved visual instruction tuning. We evaluate LLaVA-NeXT-13B as a representative open-source model for image-grounded dialogue and visual reasoning.

\textbf{DeepSeek} is represented by Janus-Pro-7B, an open-source multimodal model designed for general-purpose vision-language understanding and reasoning.

\textbf{GLM-V} is represented by GLM-4.6V, a large-scale multimodal model intended for strong image-text understanding and multimodal reasoning.

In our experiments, closed-source models are assessed via official APIs, whereas open-source models are run locally with publicly available weights and standardized configurations.

\subsection{LLM Extractor and Evaluator}
\label{appendix:llm_evaluator}
\begin{figure}[t]
    \centering
    \includegraphics[width=1\linewidth]{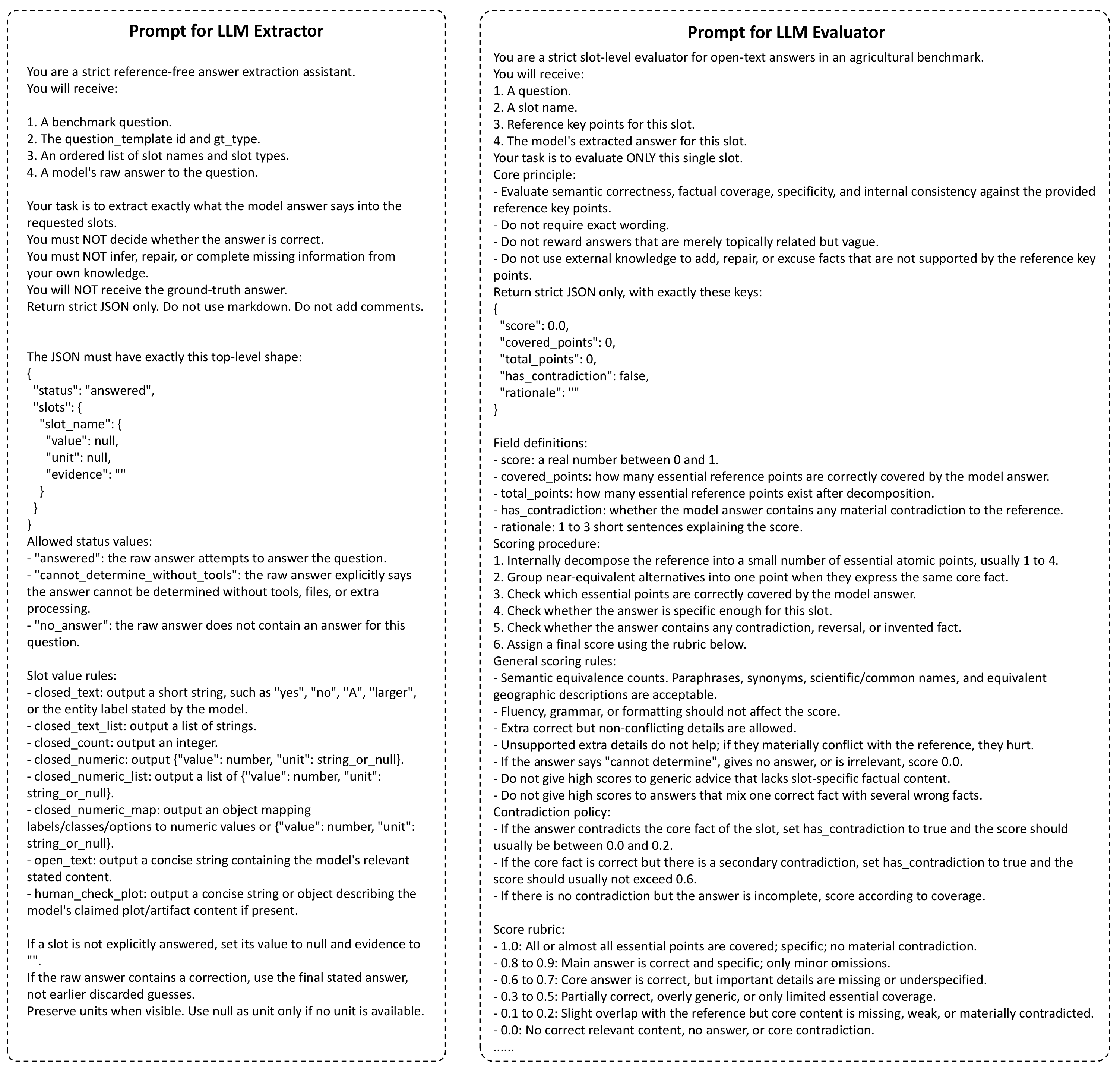}
    \caption{Prompt templates used for LLM-based extraction and evaluation}
    \label{figappendix:LLM Prompt}
\end{figure}

In this section, we show the prompt templates used for LLM-based extraction and evaluation. The full prompts, tool schemas, and parsing scripts are included in the released code. 
\paragraph{LLM Extractor.}
We employ an LLM-based extractor (GPT-4o~\cite{leon2025gpt}) solely as a reference-free normalization module that converts raw model responses into a unified slot-based representation before scoring. The extractor receives the benchmark question, the question template and answer type, an ordered list of slot names and slot types, and the raw model answer; crucially, it never sees the ground-truth answer. It is explicitly instructed to extract only what the model has stated, rather than infer, repair, or complete missing content from external knowledge. To reduce output variance, the extractor is run with deterministic decoding and required to return strict JSON with a fixed schema, including a global status field and per-slot \texttt{value}, \texttt{unit}, and \texttt{evidence} fields. If the output is unparsable or schema-inconsistent, the system automatically re-prompts the extractor with the previous error message and invalid output, allowing self-correction of formatting failures while preserving the same evidence. This design keeps the extractor narrowly scoped and auditable: it standardizes heterogeneous free-form answers without using reference answers or making any correctness judgment.
See Fig.~\ref{figappendix:LLM Prompt} for specific prompts.

\paragraph{LLM Evaluation for Open-Text Answers.}
LLM-based judging (GPT-4o~\cite{leon2025gpt}) is used only for open-text slots, whereas all closed-form slots, including categorical, counting, list-valued, and numeric fields, are scored by deterministic matching rules. Each open-text slot is evaluated independently using the question, the slot name, the reference key points for that slot, and the extractor-normalized model answer. The judge is instructed to decompose the reference into a small set of essential atomic facts, assess factual coverage, specificity, and internal consistency, and return a scalar score together with the number of covered points, the total number of reference points, an explicit contradiction flag, and a short rationale. To reduce ambiguity across heterogeneous answer types, the evaluator further appends slot-specific guidance for semantically difficult fields, such as pesticide recommendation, control method, disease confirmation rationale, and causal evidence. The judge is run with deterministic decoding and JSON-only outputs, and parsing failures trigger automatic retries; if LLM judging still fails, the evaluator falls back to a lexical-similarity score rather than silently accepting an invalid output. In our experiments, we verified that all LLM-based assessments returned valid outputs. As a result, LLM judgment is confined to cases where exact matching would be overly brittle, while the scoring process remains decomposed, traceable, and case-auditable.

\paragraph{LLM Evaluation for Plot Tasks.}
Plot-generation questions are evaluated with a dedicated two-stage LLM-based scorer (GPT-4o~\cite{leon2025gpt}) rather than the generic slot evaluator. The first stage is a visual judge that directly compares the model-generated plot image against the ground-truth figure and scores visual faithfulness, rendered data correctness, and compliance with chart-type requirements. The second stage is a parameter judge that compares the model's plotting command and upstream tool outputs against the corresponding ground-truth plotting artifacts, thereby verifying whether the underlying plotted data, axis semantics, legend semantics, and chart family are correct independently of superficial rendering style. Both stages use deterministic decoding, strict JSON outputs, and automatic retries. The final plot score is computed by averaging the visual score and the parameter score, which prevents visually plausible but semantically incorrect plots from receiving inflated credit. Missing plot images or missing plotting commands are handled conservatively by assigning zero or partial credit only when the available evidence supports it. This design makes the plot evaluator more reliable than a single holistic visual judgment, because it jointly checks both rendered appearance and plotting semantics.

\subsection{Human Participants}
We recruited six undergraduate volunteers for the human verification procedures in this study. All volunteers had relevant training in both computer science and agriculture-related topics, which enabled them to understand the benchmark tasks and domain-specific terminology with limited ambiguity. To reduce annotator-dependent variation, we selected participants with comparable academic backgrounds, used standardized written instructions and fixed rubrics, assigned non-overlapping subsets with similar workloads, and released the corresponding annotation guidelines, templates, and scoring rules in the codebase for reproducibility.

\begin{figure}
    \centering
    \includegraphics[width=1\linewidth]{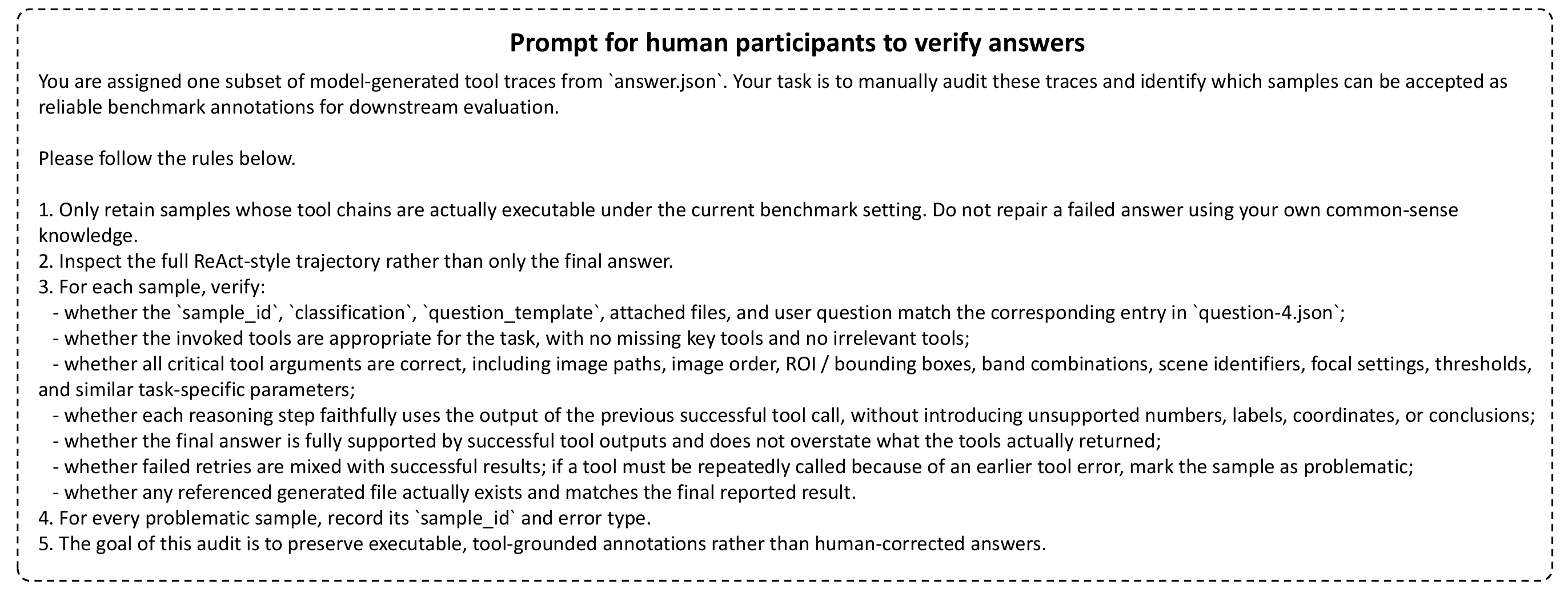}
    \caption{Answer verification prompt for human participants}
    \label{figappendix:human prompt}
\end{figure}

\paragraph{Manual Verification of Benchmark Annotations.}
Five volunteers audited the initial model-generated tool traces used in benchmark construction. The traces were partitioned into five near-equal subsets (108/108/108/108/107 samples), and each volunteer audited one non-overlapping subset. Annotators were instructed to assess the full ReAct-style trajectory rather than only the final answer, and to retain only samples whose tool chains were genuinely executable under the current benchmark setting, without repairing failed cases using external common-sense knowledge. For each sample, they verified the consistency of the metadata, question text, and associated files; the appropriateness of the invoked tools; the correctness of key tool arguments such as image paths, region specifications, band combinations, scene identifiers, focal settings, and numerical thresholds; the faithfulness of each reasoning step to the preceding tool output; the fidelity of the final answer to successful tool outputs; and the existence and consistency of any referenced generated artifacts. Problematic cases were explicitly recorded by sample ID and error type for subsequent cleanup. This protocol was intended to minimize subjective post hoc correction and to ensure that the final benchmark annotations reflect executable tool-based reasoning rather than human repair. As shown in Fig.~\ref{figappendix:human prompt}, we present the prompt provided to human participants for answer verification.

\paragraph{Human Verification of Open-Text Scores.}
We additionally conducted a targeted human audit of open-text scoring on six representative models. The evaluation materials were organized into six anonymized packages (\texttt{model1} to \texttt{model6}) and assigned to six volunteers, each responsible for one non-overlapping package. Each package contained 143 slot-level evaluation items and exposed only four fields: \texttt{question}, \texttt{slot\_name}, \texttt{reference\_answer}, and \texttt{model\_extracted\_answer}. The prompt guidelines for manual scoring are kept consistent with those for LLMs, ensuring reliable comparability, as shown in Fig.~\ref{figappendix:LLM Prompt}.

\subsection{Hyperparameters and Inference Settings}
\label{appendix:experiment details}
We build the evaluation pipeline directly with Lagent and adopt a ReAct-style interaction schema to support structured reasoning and tool invocation. In all experiments, we use the same decoding hyperparameters for all evaluated models whenever supported: temperature is set to 0.1, top-\(p\) is set to 0.7, and \texttt{max\_new\_tokens} is set to 2048. This setting is used to improve comparability across models while keeping generation relatively deterministic. For models that support an explicit reasoning mode or reasoning-depth control, such as Doubao-2.0-Pro, we follow the provider-recommended setting.

\subsection{Supplementary Experimental Results}
\paragraph{Plot-task performance.}
Table~\ref{tabappendix:plotscore} shows that plot generation remains a challenging capability even for comparatively strong multimodal models. GPT-5.4 achieves the strongest overall performance, combining the highest success rate with the best visual and parameter-level fidelity, while Doubao-Seed-2.0-Pro and Qwen3.5-35B-A3B also exhibit relatively stable behavior but with clearly lower overall scores. More broadly, the spread across models suggests that successful plot evaluation in our benchmark depends not only on producing a plot image, but also on faithfully rendering the intended data and plotting specifications.

\begin{figure}
    \centering
    \includegraphics[width=0.9\linewidth]{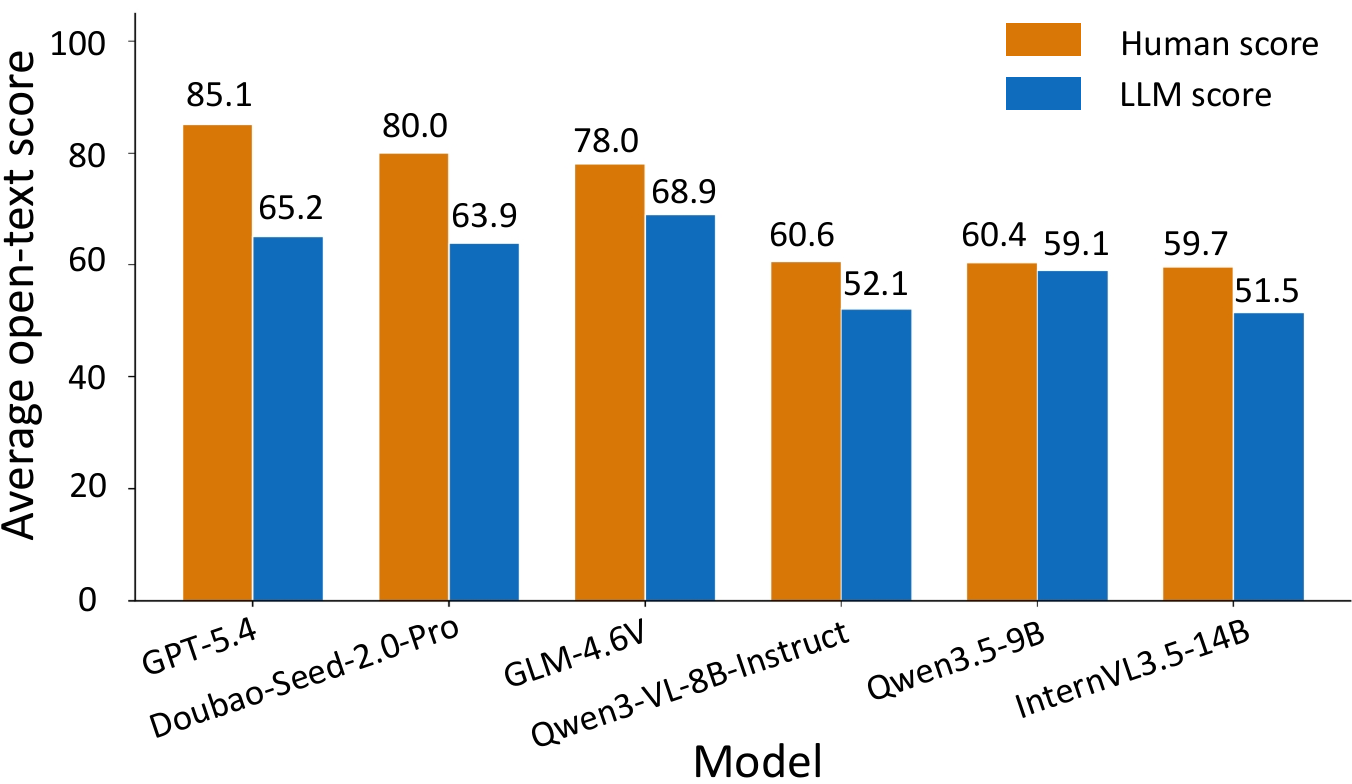}
    \caption{Human and LLM Scoring Comparison Across Different Models}
    \label{figappendix:LLMandHumanScore}
\end{figure}

\paragraph{Human verification of open-text scoring.} Fig.~\ref{figappendix:LLMandHumanScore} compares human-assigned and LLM-assigned open-text scores on six representative models over the audited subset. The two evaluations show a broadly consistent separation between stronger and weaker models: GPT-5.4, Doubao-Seed-2.0-Pro, and GLM-4.6V remain the strongest group under both protocols, while Qwen3-VL-8B-Instruct, Qwen3.5-9B, and InternVL3.5-14B receive lower scores overall. At the same time, human scores are consistently higher than the LLM-based scores, with especially large gaps on stronger models, e.g., 85.1 vs. 65.2 for GPT-5.4 and 80.0 vs. 63.9 for Doubao-Seed-2.0-Pro, whereas Qwen3.5-9B shows much closer agreement (60.4 vs. 59.1). This pattern suggests that the LLM judge is generally more conservative in scoring open-text answers, but still preserves the main relative performance trends across models, supporting its use as a scalable evaluator for open-text questions in AgroTools.

\begin{table}[t]
\centering
\caption{Summary of plot-task performance across evaluated models. Best results across all models are shown in \textbf{bold}.}
\label{tabappendix:plotscore}

\begin{adjustbox}{max width=\linewidth}
\begin{tabular}{l|c|c|c|c}
\toprule
\textbf{Model} 
& \textbf{Plot Success Rate} 
& \textbf{Plot Visual Score} 
& \textbf{Plot Parameter Score} 
& \textbf{Overall Plot Score} \\
\midrule
GPT-5.4 & \textbf{0.808} & \textbf{0.715} & \textbf{0.789} & \textbf{0.752} \\
Doubao-Seed-2.0-Pro & 0.596 & 0.577 & 0.588 & 0.583 \\
Qwen3.5-35B-A3B & 0.558 & 0.431 & 0.434 & 0.432 \\
InternVL3.5-8B & 0.308 & 0.217 & 0.258 & 0.237 \\
Qwen3-VL-8B-Instruct & 0.596 & 0.230 & 0.162 & 0.196 \\
GLM-4.6V & 0.385 & 0.184 & 0.129 & 0.157 \\
Qwen3.5-9B & 0.135 & 0.118 & 0.117 & 0.118 \\
InternVL3.5-14B & 0.192 & 0.078 & 0.063 & 0.071 \\
Gemini-2.5-Pro & 0.211 & 0.317 & 0.268 & 0.293 \\
\bottomrule
\end{tabular}
\end{adjustbox}

\vspace{-0.6em}
\end{table}

\begin{figure}[t]
    \centering
    \includegraphics[width=0.9\linewidth]{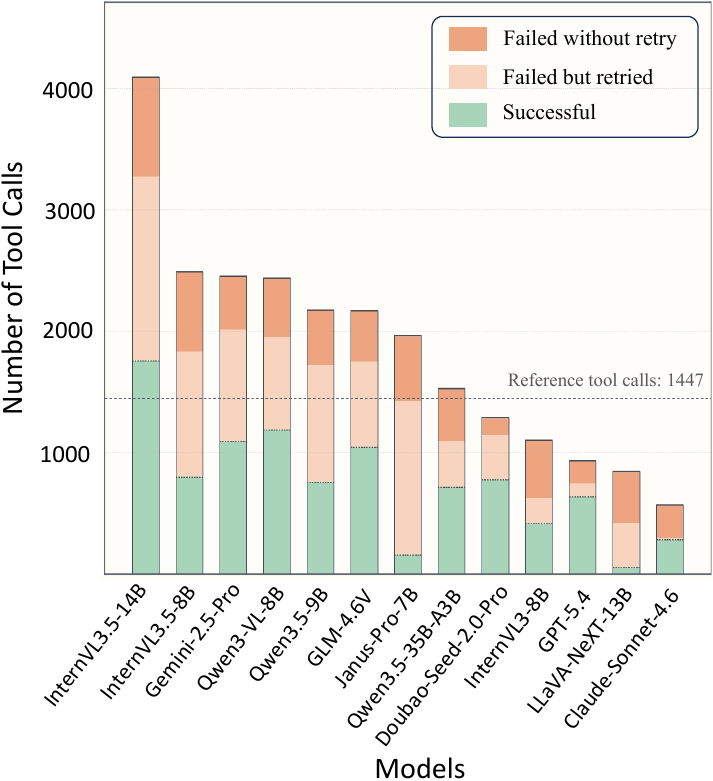}
    \caption{Tool-call outcomes and retry behavior in end-to-end evaluation.}
    \label{fig:tool error}
\end{figure}

\section{Further Discussion and Limitations}
\paragraph{Process matters.} Our experiments show that agricultural tool-use evaluation cannot be reduced to final-answer correctness alone. Many failures occur earlier in the pipeline, including incorrect tool planning, invalid argument generation, premature stopping, and weak evidence-grounded synthesis. Because agricultural tasks are often precision-sensitive, such process errors can lead to materially wrong conclusions even when the final answer appears plausible. AgroTools is therefore intended not only as a harder multimodal benchmark, but also as a diagnostic testbed for studying how reliably models interact with agricultural tools.

\paragraph{Tools are not free gains.} Tool augmentation helps strong models more consistently, but does not automatically improve weaker ones. Models with better long-context understanding and instruction following are more able to benefit from external tools, while weaker models often struggle with the added interaction format and longer prompts. This suggests that future agricultural agents need stronger tool-aware training, better action formatting, and more robust execution recovery, rather than relying on tool access alone.

\paragraph{Limitations.} To keep the benchmark executable and reproducible, AgroTools currently focuses on single-turn, image-grounded tasks with relatively short toolchains and a fixed tool environment. Real agricultural deployments may involve longer workflows, richer user interaction, temporal records, external databases, and broader sensing modalities. In addition, while AgroTools covers diverse public datasets and task families, future versions can further expand regional coverage, crop diversity, multilingual interaction, and human verification. We view these as natural extensions of the benchmark rather than contradictions to its current scope.

\section{Ethical Considerations}
The AgroTools benchmark is constructed from publicly available agricultural datasets and other publicly accessible resources. All externally sourced data is obtained from published academic repositories and used in full compliance with the original datasets’ copyright notices, licensing terms, and usage conditions. For datasets subject to more restrictive licensing or redistribution terms, we will not directly redistribute the raw data; instead, we will provide official download links together with reproducible scripts for data acquisition and preprocessing, so as to ensure strict adherence to the original licenses and terms of use. We commit to releasing the AgroTools benchmark as an open-source project, with the goal of advancing research in tool-augmented multimodal agricultural AI and fostering reproducible evaluation and iterative model development.

\section{Case Study}
\label{appendix:case study}
\begin{figure}[ht]
    \centering
    \includegraphics[width=0.8\linewidth]{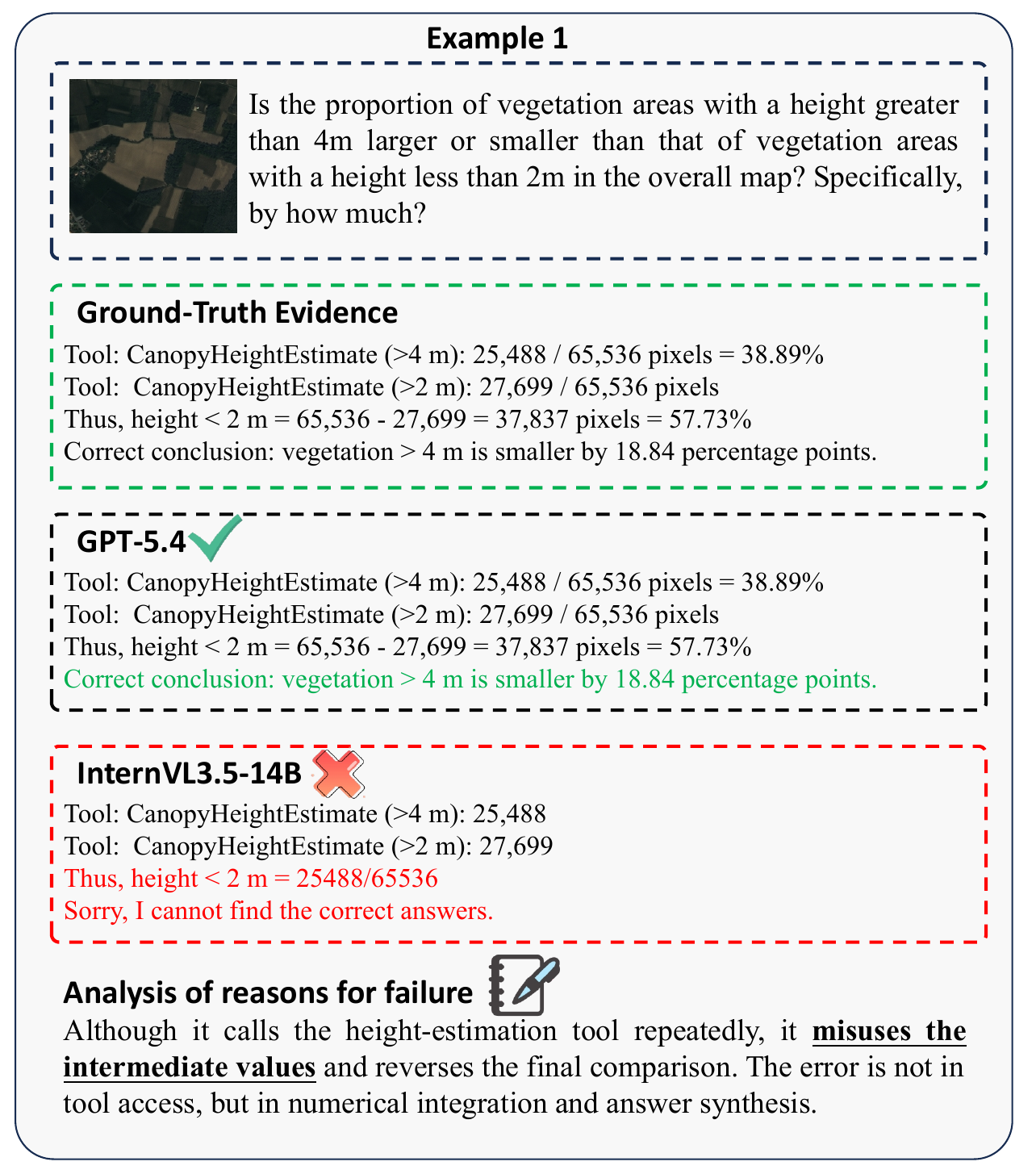}
    \caption{Case Study of GPT-5.4 and InternVL3.5-14B on AgroTools Tasks.}
    \label{fig:case1}
\end{figure}

\begin{figure}[t]
    \centering
    \includegraphics[width=1\linewidth]{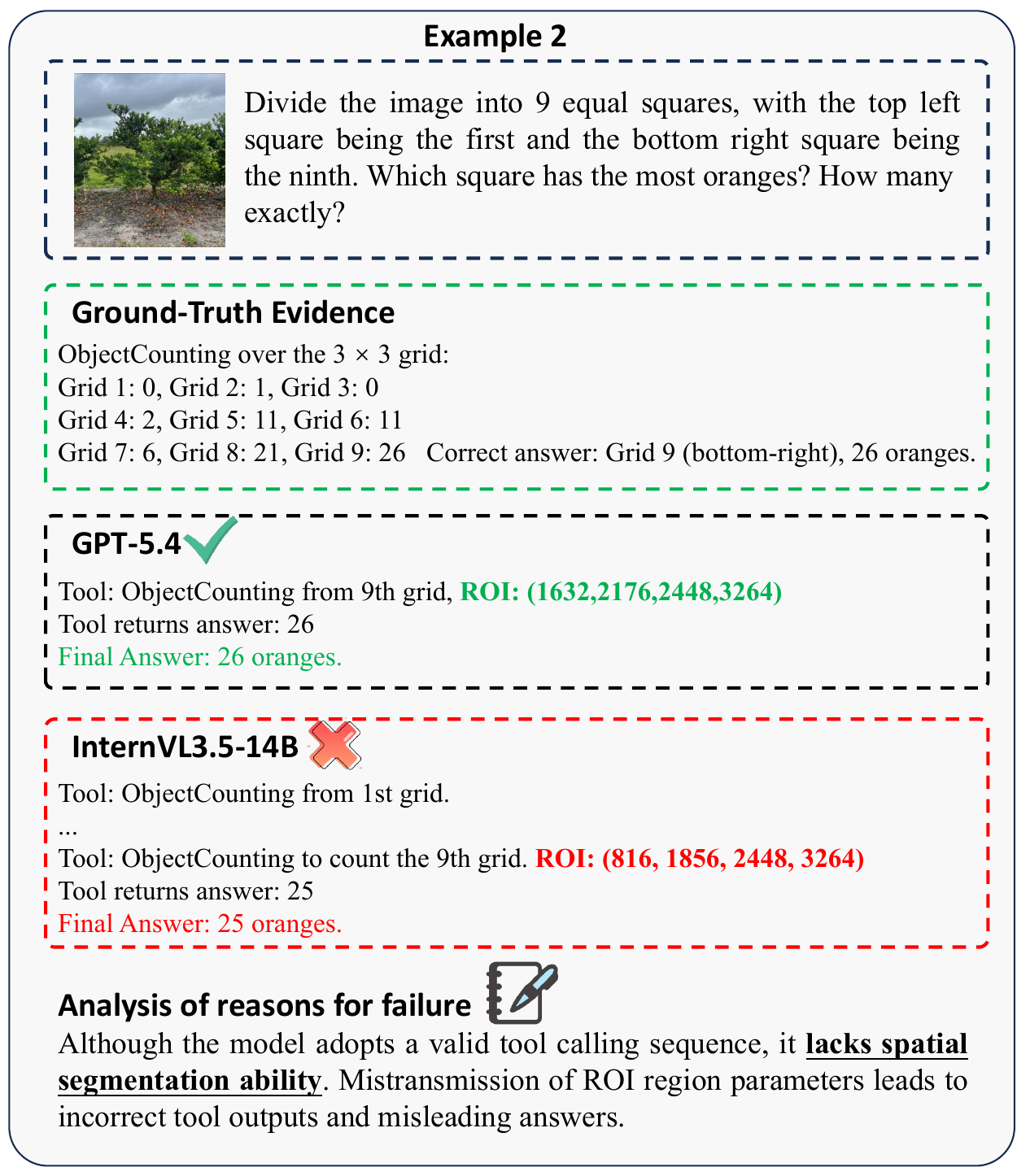}
    \caption{Case Study of GPT-5.4 and InternVL3.5-14B on AgroTools Tasks.}
    \label{fig:case2}
\end{figure}

\begin{figure}[t]
    \centering
    \includegraphics[width=1\linewidth]{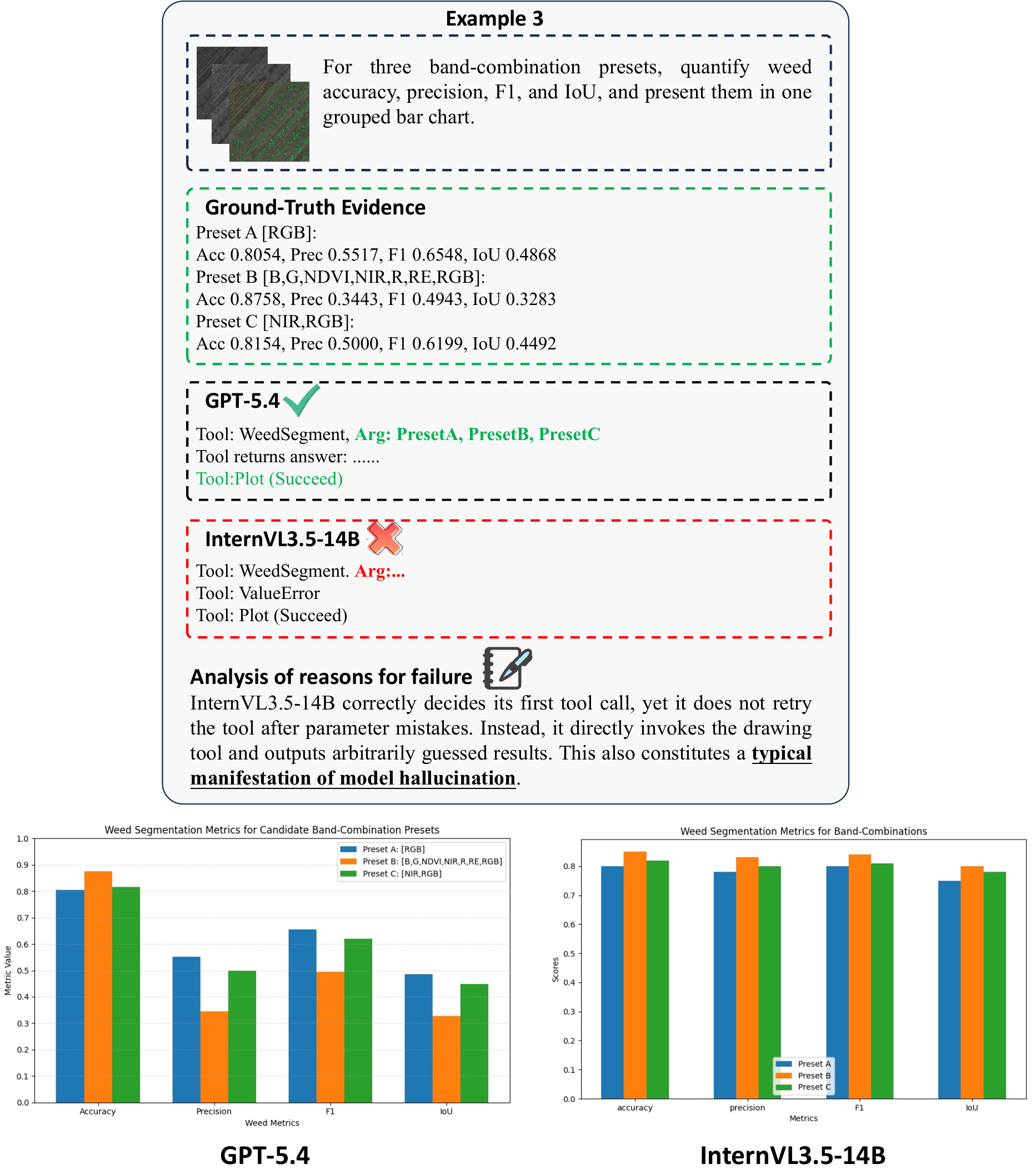}
    \caption{Case Study of GPT-5.4 and InternVL3.5-14B on AgroTools Tasks.}
    \label{fig:case3}
\end{figure}

\begin{figure}[t]
    \centering
    \includegraphics[width=1\linewidth]{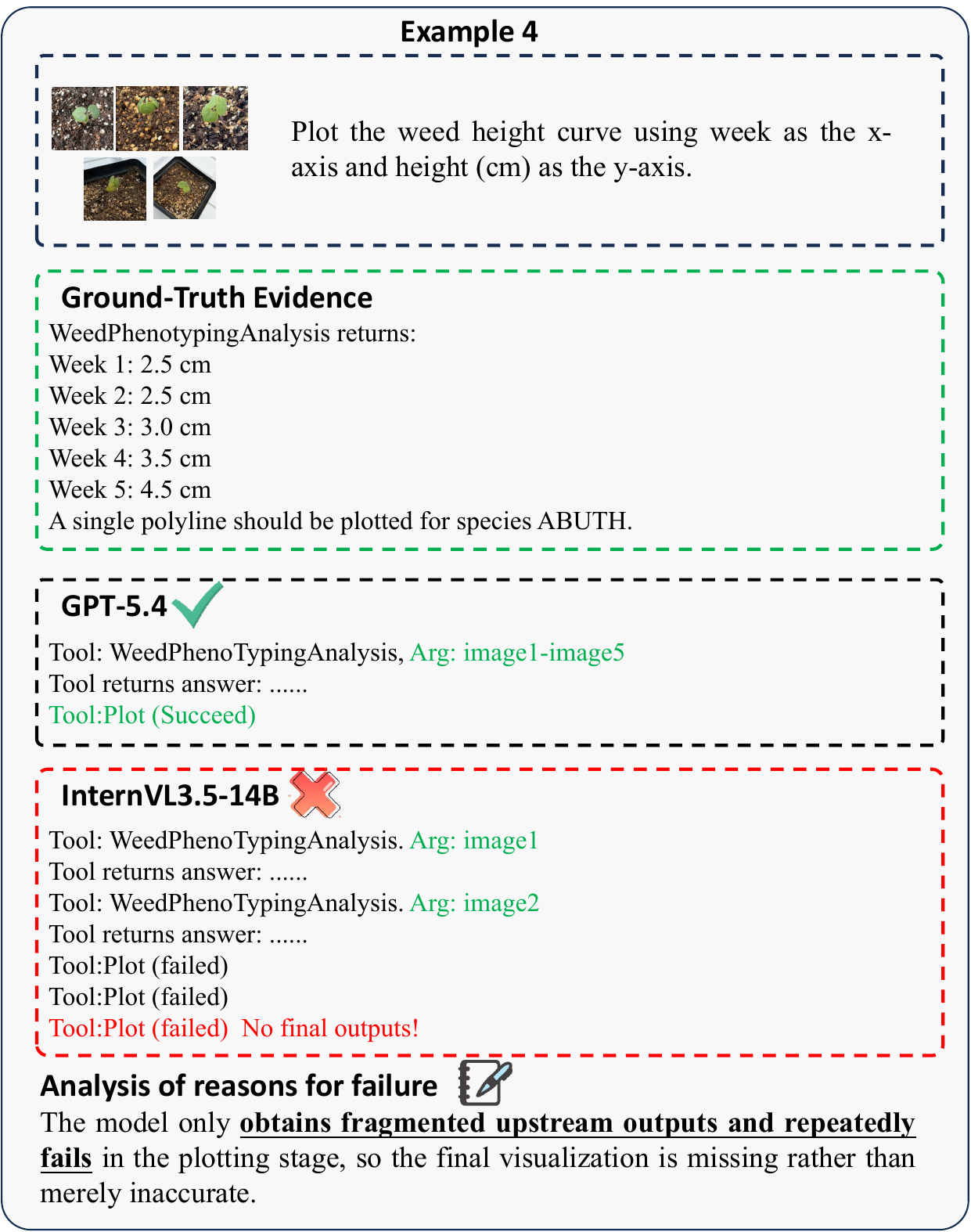}
    \caption{Case Study of GPT-5.4 and InternVL3.5-14B on AgroTools Tasks.}
    \label{fig:case4}
\end{figure}

\begin{figure}[t]
    \centering
    \includegraphics[width=1\linewidth]{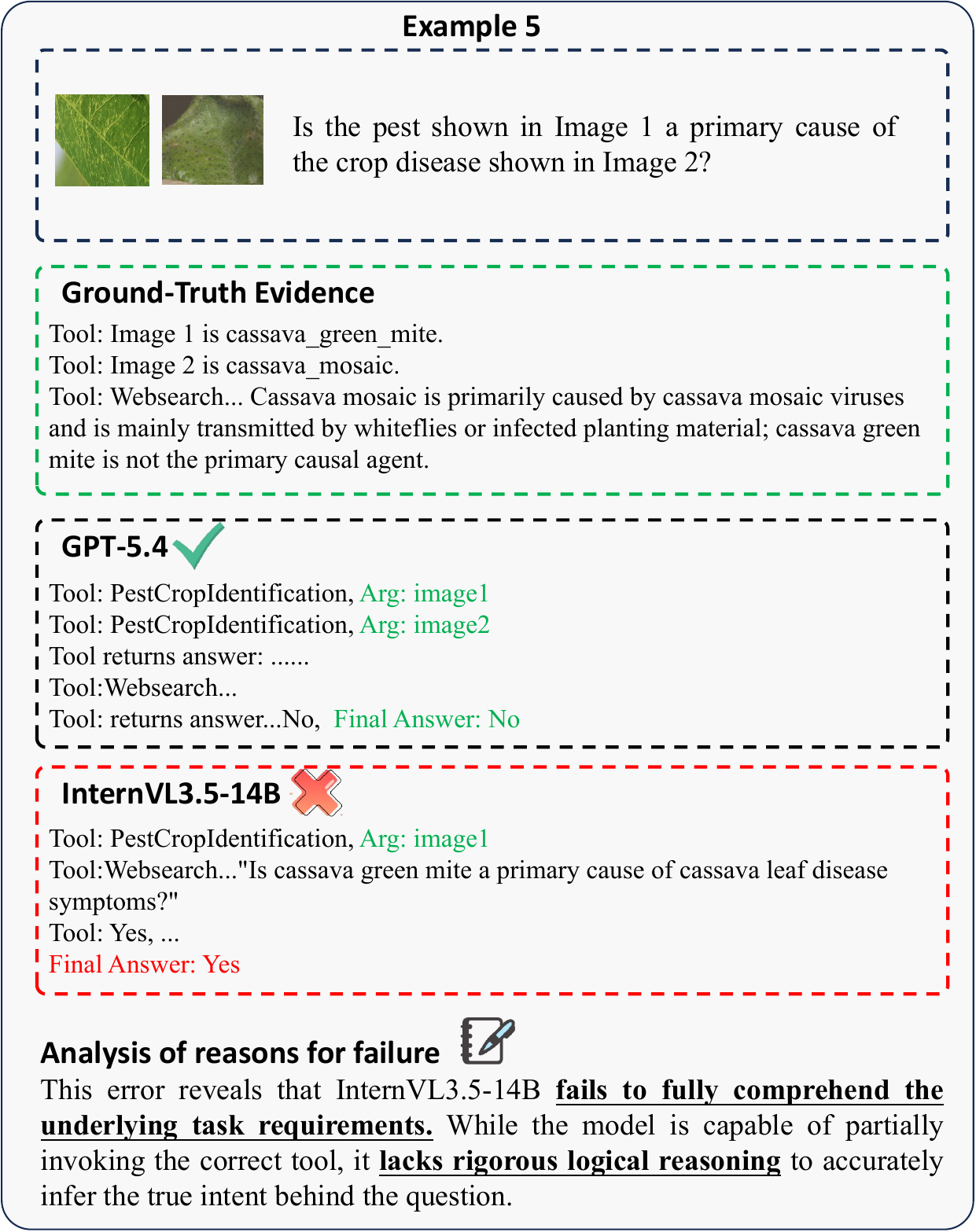}
    \caption{Case Study of GPT-5.4 and InternVL3.5-14B on AgroTools Tasks.}
    \label{fig:case5}
\end{figure}

\begin{figure}[t]
    \centering
    \includegraphics[width=1\linewidth]{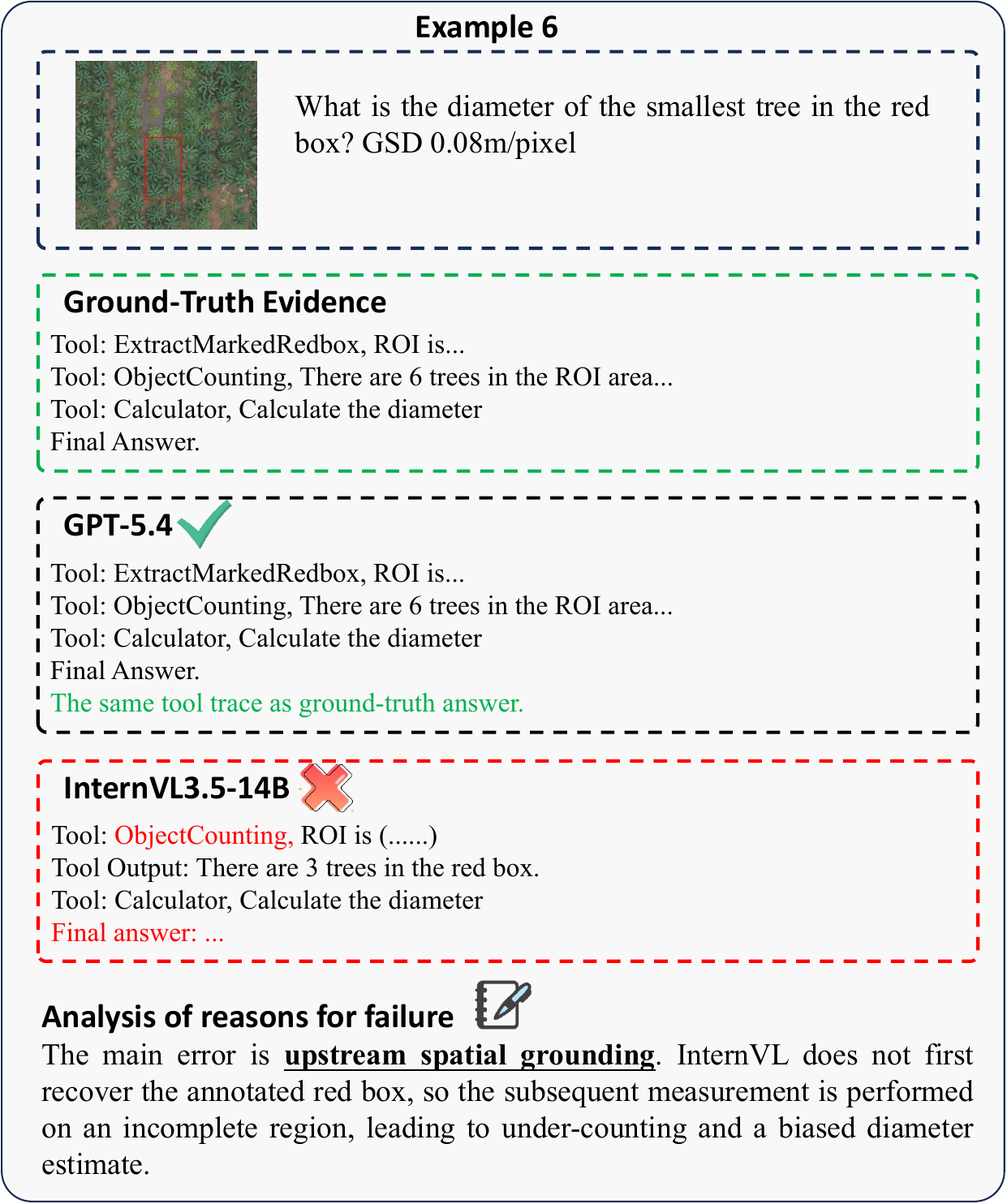}
    \caption{Case Study of GPT-5.4 and InternVL3.5-14B on AgroTools Tasks.}
    \label{fig:case6}
\end{figure}

\begin{figure}[t]
    \centering
    \includegraphics[width=1\linewidth]{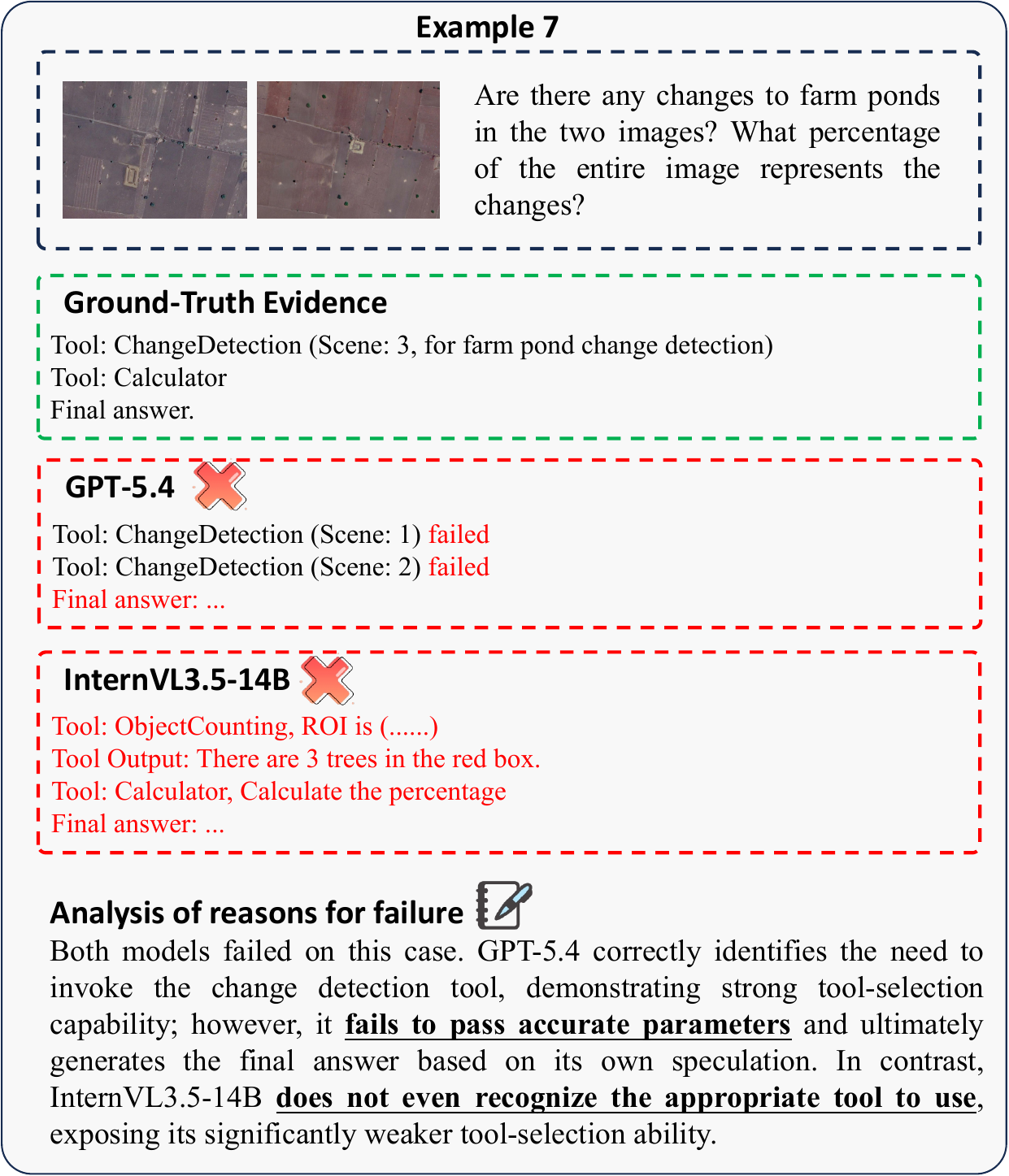}
    \caption{Case Study of GPT-5.4 and InternVL3.5-14B on AgroTools Tasks.}
    \label{fig:case7}
\end{figure}

In this section, we present representative question cases from different AgroTools tasks and the corresponding responses generated by GPT-5.4 and InternVL3.5-14B (see Fig.~\ref{fig:case1} - Fig.~\ref{fig:case7}).
By comparing these two models, we aim to provide insights into both the upper performance bound of current tool-augmented LMMs and the capabilities of open-source alternatives in agricultural scenarios. The cases cover diverse task dimensions from AgroTools. Each case consists of a visual input, a task-specific question, and the model interaction process, including tool calls, parameter passing, and final answers.
Through these examples, we observe that GPT-5.4 consistently demonstrates strong tool-use proficiency, including accurate tool selection, correct parameter specification, and effective synthesis of tool outputs into coherent answers, even in complex multi-step tasks. In contrast, InternVL3.5-14B, while generally capable of identifying the need for tool invocation, frequently struggles with parameter correctness. In some cases, it even fabricates answers without relying on tool results. These differences highlight the varying levels of tool-use robustness, reasoning capability, and domain adaptation in current tool-augmented LMMs.

\end{document}